\pdfoutput=1

\documentclass[11pt]{article}

\usepackage{acl}

\usepackage{times}
\usepackage{latexsym}

\usepackage{subcaption}

\usepackage{multicol,multirow}

\usepackage[T1]{fontenc}

\usepackage[utf8]{inputenc}

\usepackage{microtype}

\usepackage{inconsolata}

\usepackage[shortlabels,inline]{enumitem}
\usepackage{cleveref}
\crefname{section}{s}{Sections}
\crefname{table}{Table}{Tables}
\crefname{figure}{Fig.}{Figs.}
\crefname{algorithm}{Alg.}{}
\crefname{ALC@unique}{Line}{Lines}
\crefname{equation}{Eq.}{Eqns.}
\crefname{appendix}{Appendix}{}
\crefformat{section}{\S#2#1#3}
\usepackage[normalem]{ulem}
\usepackage{graphicx}

\newcommand{\movedtoappendix}[1]{}

\title{Examining Participant-specific Goals for a Deeper Understanding of Complex Events}
\title{SAGA: Story Alternatives and Examining Participant-specific Goal Applicability for a Deeper Understanding of Complex Events}
\title{SAGA: A Participant-specific Examination of Story Alternatives and Goal Applicability for a Deeper Understanding of Complex Events}

\author{ %
Sai Vallurupalli$^1$,\ \ %
Katrin Erk$^2$,\ \ %
Francis Ferraro$^1$
\\
$^1$ University of Maryland, Baltimore County, \\
$^2$ University of Texas, Austin \\
  \texttt{kolli@umbc.edu},
  \texttt{katrin.erk@utexas.edu},
  \texttt{ferraro@umbc.edu}\\
  }

\begin{document}
\maketitle
 
\begin{abstract}
\label{sec:abstract}

Interpreting and assessing goal driven actions is vital to understanding and reasoning over complex events. It is important to be able to acquire the knowledge needed for this understanding, though doing so is challenging. %
We argue that such knowledge can be elicited through a \textit{participant achievement} lens. We analyze a complex event in a narrative according to the intended achievements of the participants in that narrative, the likely future actions of the participants, and the likelihood of goal success. %
We collect 6.3K high quality goal and action annotations reflecting our proposed participant achievement lens, with an average weighted Fleiss-Kappa IAA of 80\%. Our collection contains annotated alternate versions of each narrative. These alternate versions vary minimally from the ``original'' story, but can license drastically different inferences. %
Our findings suggest that while modern large language models can reflect some of the goal-based knowledge we study, they find it challenging to fully capture the design and intent behind concerted actions, even when the model pretraining included the data from which we extracted the goal knowledge. %
We show that smaller models fine-tuned on our dataset can achieve performance surpassing larger models. %
 

\end{abstract}

\section{Introduction}
\label{sec:intro}

\begin{figure}[t]
    \centering     
    \includegraphics[trim=0.25in .25in 0in 0in, scale=.45]
    {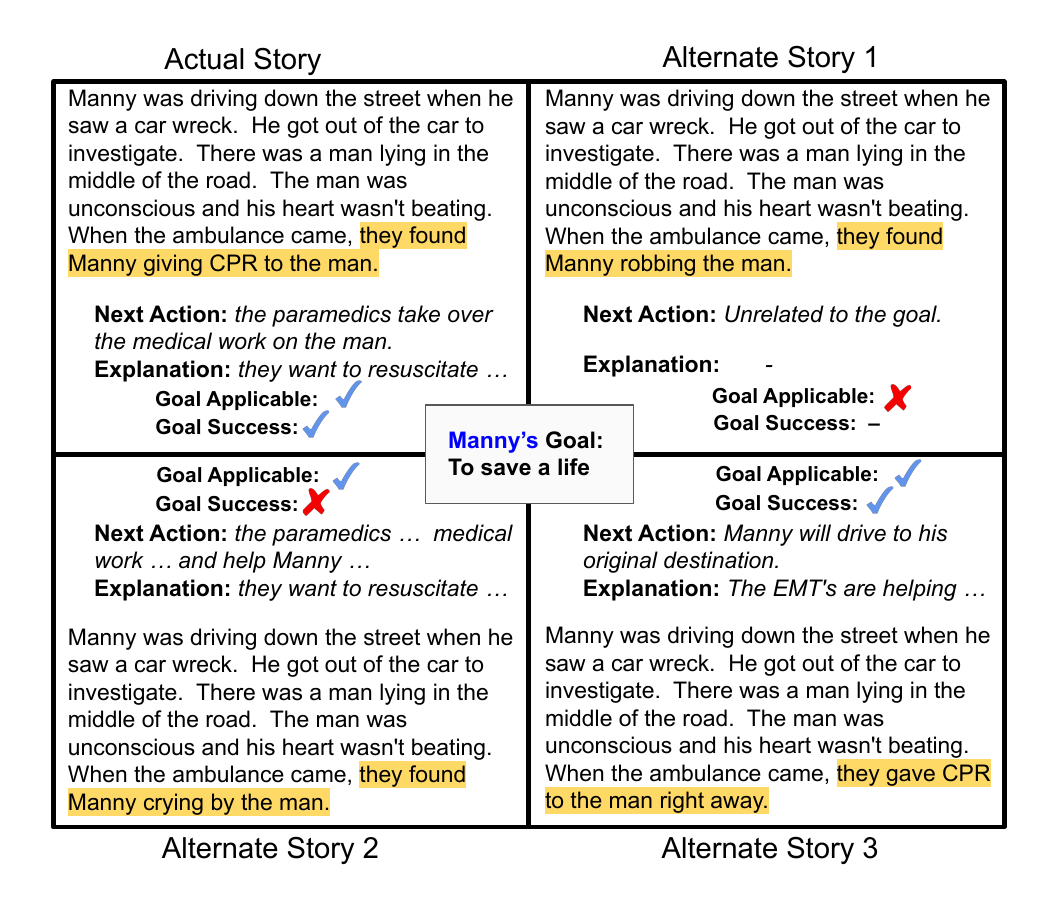}
	\caption{A participant's goal inferred from the actual story when applied to 3 alternative stories, drawn from the PASTA dataset~\cite{ghosh2023pasta}; slightly varying actions in the stories lead to different goal achievement outcomes. %
 }
	\label{fig:fig1}
 \vspace{-5mm}
\end{figure}

Understanding \textit{goals} is central to human understanding of text~\citep{foss_and_bower_1986}. However, this is challenging, as slight variations in actions reported in text can lead to vastly different goals, achievement outcomes and future actions. 
Consider the stories in \cref{fig:fig1}, 
where the actions of Manny (the \textit{participant}) in the ``original'' story (top left) indicate his goal of saving a life, which he is able to achieve at the end of the story. However, small changes in the narrative can lead to vastly different inferences. %
Consider the three alternative stories shown: in alternative 1 (top right), a different action points to Manny having a different goal. On the other hand, while the different actions in the other two alternatives leave Manny's goal unchanged, future actions and goal achievement are different. %
In this work, we \textbf{simplify the complex task of understanding and reasoning about a participant's goal in a narrative by decomposing it into the actions, intentions, and plans} that the participant takes or may make in the future. %

\begin{figure*}[t]
    \centering
\includegraphics[trim=0.2in .4in 0in 0in, scale=.43] 
    {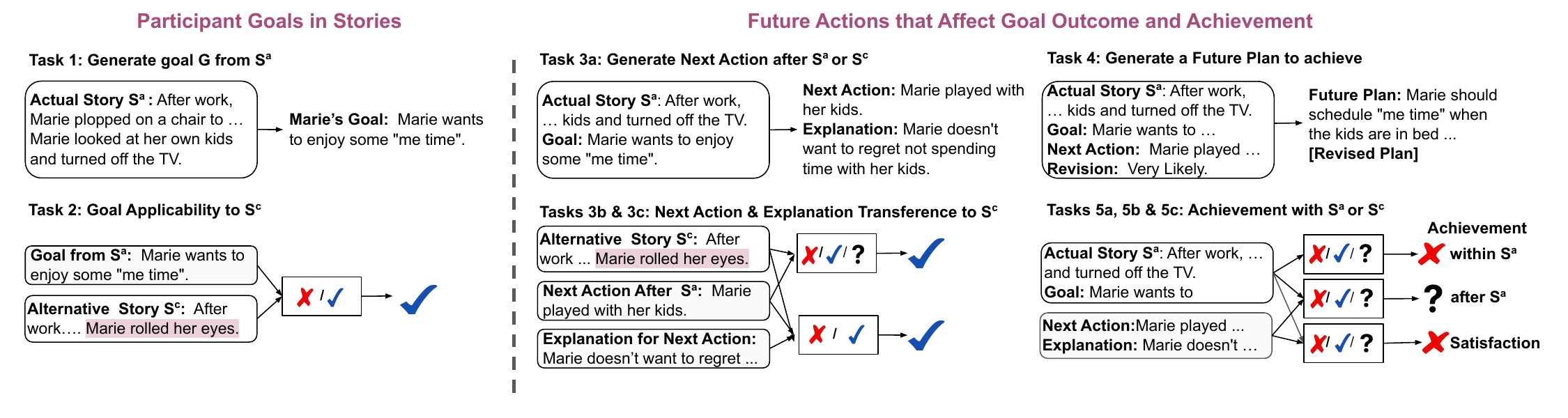}
	\caption{Goal reasoning inferences from our dataset formulated as benchmarking tasks.  These consist of  both generating a  participant's goal and future actions after the story aimed at goal achievement and identifying goal applicability and achievement.  Tasks 1, 3a and 4 examine the generative understanding of goals, explainable future actions and plans. Tasks 2, 3b, 3c and 5 examine  discriminative understanding of applicability and achievement.  
 }

	\label{fig:fig2}
 \vspace{-3mm}
 \end{figure*}

While knowledge of participants' intentions and how they are impacted by situations is strategic for communication and language understanding~\citep{zhang2023ask,ammanabrolu-etal-2021-motivate,callison-burch-etal-2022-dungeons,cao-etal-2022-goal}, acquiring the necessary goal and plan knowledge is not  straightforward. %
First, situations in narrative text can be very complex! This is partly due to the implied nature of described situations~\cite{poque}, the subjective nature of someone's understanding of a participant's goal~\cite{graesser,foss_and_bower_1986}, and our ability to understand a goal at varying levels of abstraction and specificity. %
Second, subjectivity and incomplete information often require \textit{prospective reasoning}
to identify the most likely outcome in the immediate future, inferring information that is unstated yet assumed~\cite{hovy-yang-2021-importance,Davis2015CommonsenseRA}.  We also must be able to distinguish reasonable vs. unreasonable alternatives~\cite{niven-kao-2019-probing,zellers2018}. Both of these are challenging. 
Third, while pre-trained large language models (LLMs) are powerful~\citep[\textit{inter alia}]{brown,wei2022finetuned}, they may perform poorly on tasks requiring robust reasoning~\cite{ghosh2023pasta,Zellers2019HellaSwagCA,qin-etal-2019-counterfactual}. %

Our formulation ultimately supports inferences of the form ``who achieved what and under what conditions,'' i.e., inferences useful in application areas that rely on broader reasoning, like claim verification or question answering. %
We consider the rationale for ``why'' a participant carries out certain actions (\textit{goal knowledge}) and their plans to achieve their goal by considering ``what-next'' type of future actions and plans (\textit{prospective knowledge}). 
To aid our study and future work, we use our framework to analyze alternative stories and how small variations in the narrative affect what inferences can be made. %
Via a multi-stage annotation pipeline, we collected 6225 goal annotation sets (consisting of 886 ``actual'' \& 951 ``alternative'' stories). %
From this rich data, we formulate multiple inference tasks of increasing difficulty, seen in \cref{fig:fig2}, including what can be inferred about the goals of participants (\textit{Tasks 1 \& 2}), future actions that affect the goal outcome (\textit{Tasks  3 \& 4}) and achievement (\textit{Task 5}). %

We examine LLMs on these tasks; since models trained with human feedback have shown impressive performance on understanding human intent~\cite{stiennon,ouyang2022training}, we compare GPT versions 4, 3.5-Turbo, and various sizes of the T5 family of models Flan-T5 and T5 models. 
 We use prompting and fine-tune models on our collected dataset to examine the differences with both both options and the benefits of fine-tuning smaller models. Our results and analysis show the strengths and weaknesses of these various models. 
We find that pre-trained larger models generally perform better than smaller models and contain less factual errors. %
Few-shot prompting is especially useful for tasks requiring within-story details.  Some of the stories that are part of our dataset are part of the pretraining data for Flan-T5~\cite{wang-etal-2022-super, longpre2023flan}; we examine how models handle subtle changes in these stories and find that  few-shot prompting and fine-tuning help correct these errors.   %
Overall, we find that even the larger models struggle to hone-in on the human intent behind a set of actions.  Fine-tuning and few-shot prompting improve smaller models to be competitive with, or surpass, the larger models.    

We summarize our contributions as follows: 
\begin{enumerate*}[(1)]
\item We construct a dataset, \textbf{SAGA} (\textbf{S}tory \textbf{A}lternatives and \textbf{G}oal \textbf{A}pplicability), of overarching goals from alternative stories to help gain a broader and deeper understanding of complex events through goal-based reasoning. 
\item We developed a multi-tier pipeline that allows crowd workers to provide subjective judgements and free-form text annotations for different story versions in several progressive stages.
\item We leveraged cross-disciplinary narrative understanding research in psychology, philosophy and linguistics to inform and develop streamlined annotation and evaluation processes to collect a high quality dataset.
\item We formulated important inferences that can be made from our dataset, designed challenge tasks with these inferences and benchmarked several intent based LLMs, 
demonstrating areas where they under-perform humans. %
Our data and code are available at \url{https://github.com/saiumbc/SAGA}. %

\end{enumerate*}

\section{Related Work}
\label{sec:related}
\paragraph{Goals:}
Cross-disciplinary research in psychology argues that super-ordinate goals disentangle the temporal order from the discourse order of their actions~\cite{reasoned_action,graesser} and that cognitive modeling of the goal oriented actions can seamlessly combine situational and commonsense knowledge leading to a deeper understanding of situations
~\cite{graesser_how,zacks_event,carpendale_2015}. 

Natural Language Understanding explored goal oriented reasoning through activities in first person narratives~\cite{rahimtoroghi-etal-2017-modelling}, location-based actions in news-text ~\cite{jiang-riloff-2018-learning}, chaining and ordering of action plans from procedural text \cite{zhang-etal-2020-reasoning} and schema construction \cite{lyu-etal-2021-goal}. 
Leveraging participants' goals, recently, \newcite{bellos-etal-2024-large}  examined LLMs' sequential reasoning capabilities. These works explore the procedural and sequential relationship between activities, plans and explicit goals.  Another line of research,
commonsense based implicit question answering \cite{lal-etal-2022-using,lal-etal-2021-tellmewhy,10.1162/tacl_a_00370} collected goals as answers; the goals are implied by the context but their level of abstraction is question-dependent.  We collect implied goals at the highest level of abstraction from the actions and intentions of a volitional participant and from varied alternative situations.   

\paragraph{Inferences in Alternative Situations:}
 Proposing the common sense natural language inference task, SWAG~\cite{zellers2018} and HellaSWAG~\cite{Zellers2019HellaSwagCA}, explored sentence completion in alternative situations.   While these tasks examined counterfactual reasoning in discriminative models, TIMETRAVEL~\cite{qin-etal-2019-counterfactual} studied counterfactual knowledge possessed by generative models for alternative story rewriting.   PASTA~\cite{ghosh2023pasta}  introduced implied states that stories depend upon and examined the tasks of state inference from alternative stories and story rewriting for alternative states.  We use the alternative stories from PASTA to examine changes in goal inference in alternative situations. \newcite{storks-etal-2021-tiered-reasoning, jiang} used alternative stories to examine situation plausibility through procedurally tracking natural physical laws or participants' physical states; we examine goal plausibility through participants' intentional actions.

\section{Decomposing Goal Understanding}
\label{sec:def}

Reasoning over goals has natural subjectivity and complexity. %
We handle this in a few ways. %
First, we borrow from cross-disciplinary research ~\cite{Goal_Constructs_in_psychology,gollwitzer_and_bargh} to reduce the subjectivity in goals by collecting overarching \textit{\bfseries super-ordinate} goals: a super-ordinate goal is a complex higher order goal that is achieved through several actions. This super-ordinate goal licenses the rationale behind all of a participant's actions in a situation. %
Second, we focus on \textit{\bfseries intentions}, which can be viewed as a representation of planned actions that lead to achieving a goal~\cite{locke_2002,gollwitzer_intention,locke_1981}. 
Third, we capitalize on prior research that people are primed to think of future details for a goal that is yet to be accomplished~\cite{zacks_event,harmon,KEEFE1993446} and predict future events that help in assessing goal achievement and obtain plans (revised as necessary and when possible) that lead to achievement. %

\subsection{Stories and Participants}
\begin{figure}[t]
 \centering
  \includegraphics[trim=.6in .1in 0in 0in, scale=.6]
    {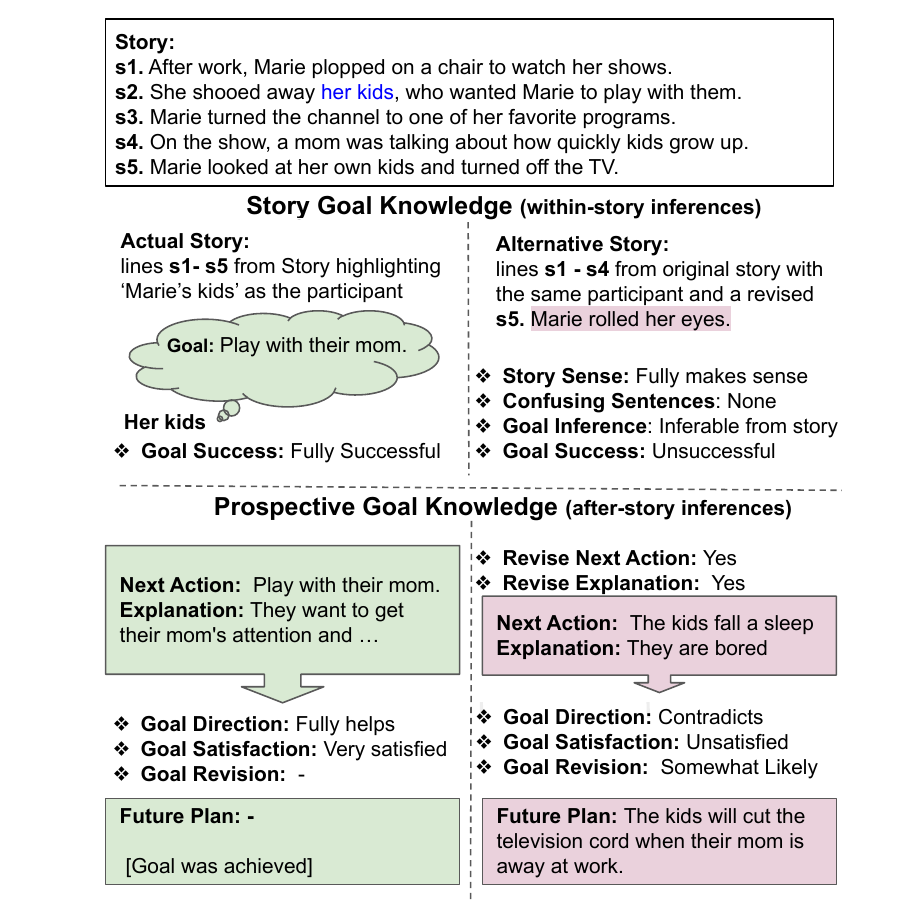}
	\caption{Our modeling of goals for ``{\color{blue}the kids}.'' In several stages a highlighted participant-specific story is annotated starting with the actual story (left) and an alternate story (right) resulting in a goal annotation set consisting of both free-form text and label assignments.%
 }
 \vspace{-3mm}
 \label{fig:fig3}
\end{figure}

We select  ROCStories~\cite{Mostafazadeh2016ACA} and their corresponding alternative stories from the PASTA dataset. %
In PASTA~\cite{ghosh2023pasta}, original ROCStories (``actual stories'') have up to three ``alternative'' stories.\footnote{ %
Each instance from PASTA has 4 components$:$ a story, a participant-state  supported by the story, an opposing state value (a ``counterfactual'' state), and an alternative story supporting this opposite state. %
The original ROCStories were annotated with pairs of contrasting states, such that one of the states is supported by the story and the other is supported by a minimally revised counterfactual (``alternative'') story. %
Also available is a list of story sentences in the original story that support the participant state. %
While this paper only uses the stories, rather than the states, the inherent link between those states and the novel goal and intentionality annotations we collect provide exciting avenues for future research. %
}
In \cref{fig:fig3} we contrast an actual vs. alternative story. %

Each actual story $S^a$ has five sentences, $\{S^a_1, \ldots, S^a_5\}$. %
As a story can include multiple participants, we first limit our analysis to participants that can act volitionally~\cite{BINSWANGER1991154}.  
We then highlight mentions of a participant $P_i$ in each sentence, and instruct annotators to infer the intentions of the highlighted participant. 

Starting with an actual story then followed by an alternate story, for each participant, we obtain $J=3$ goal annotation sets based on within-story (\cref{sec:story-goal-knowledge}) and after-story knowledge (\cref{sec:prospective-goal-knowledge}). %
Unless otherwise specified, our 5-point Likert scales range from ``most'' to ``least'' (with a middle uncertain option). %
In \cref{sec:dataeval} we discuss evaluating these annotations. We provide a detailed list of these annotations with examples in the appendix (\cref{tab:HITS-annotation}).

\subsection{Story Goal Knowledge}
\label{sec:story-goal-knowledge}
We obtain a ``super-ordinate goal'' (a goal achieved through several actions) of a participant by asking for the aim of the participant's actions, i.e., the rationale for ``why'' the actions are performed.  
To prevent comprehender-bias of selecting an immediate unfinished goal at the end of a story~\cite{trabasso}, we ask annotators to provide an overarching \textbf{goal description} $G_{ij}$ that is supported by as many of $P_i$'s actions as possible. For this goal, we also obtain \textbf{goal success} at the end of the story via a 5 point Likert scale. 


\subsection{Prospective Goal Knowledge}
\label{sec:prospective-goal-knowledge}
People can use situational context to infer future actions, especially for goals that have not yet been completed, helping assess whether the goal is achievable~\cite{zacks_event,harmon,KEEFE1993446}. Thus we ask annotators to predict and describe a likely \textbf{next action} involving the participant that follows the story. To ensure the next action is logical, we ask the annotators to provide an \textbf{explanation} for the action appropriateness. 

The next action helps determine the \textbf{goal (achievement) direction}, i.e., whether the goal is achievable or not with the action, via a 5 point Likert scale. %
When a goal is not achieved after the next action, more actions in the form of a \textbf{Future Plan} might help achieve it. 
We ask annotators to determine if a revision to the original plan observed from the story can help with goal achievement.  We obtain an appropriate original or revised plan unless the annotator thinks the goal is unachievable.  

While the next-action is not part of the story, it is highly likely to be logically continuous with the story, and so provides an appropriate point to consider whether the participant would be satisfied with their progress. 
We collect the \textbf{goal satisfaction} annotation because, despite the expectation that a participant would be satisfied with their goal achievement, it is possible the goal and the situational context could cause the participant to be unsatisfied (likewise satisfied with an unachieved goal) and identifying such goals is useful. 

\subsection{Goals in Alternative Situations}
We expect the alternative stories $S^{c_k}$ ($k \le 3$) from PASTA to alter goal inferences seen in an actual story $S^a$. A participant's actions in  $S^{c_k}$ could indicate that their goal is different from their goal in $S^a$, or that their goal is unchanged from  $S^a$ albeit with an altered course of goal achievement. To capture these variations, we identify whether the participant $P_i$ in $S^{c_k}$ intends to achieve the goal $G_{ij}$ (annotated for $S^a$) and achieves them in $S^{c_k}$.  

We notice that revisions made to obtain an alternative story can lead to missing context for the specific goals we want to reason about. %
This increases the difficulty in reasoning and, given the subjective nature of story understanding, can lead to diverging goal inferences and annotations. To control for this, we obtain annotations from 3 workers and take a majority vote (see \cref{tab:evaluation-IAA} in the appendix for IAA). Our goal inference annotations as shown in \cref{fig:fig3} (in the right column) include grading $S^{c_k}$ on its overall coherence (obtaining a list of contributing sentences when a story is confusing). %
When story coherence allows goal reasoning, we obtain annotations for whether $P_i$ intends to and achieves the goal $G_{ij}$ in $S^{c_k}$. %

For obtaining a complete goal annotation set, for each intended goal we also collect prospective goal knowledge using the $S^{c_k}$ context but use a single annotator as was done for the actual story.  Since  $S^{c_k}$ was obtained from $S_a$ with minimal revisions, when possible, we encourage  minimal revisions to the Next Action and Explanation by providing these values from $S_a$ and allowing the annotator to update them as needed.  

\section{Dataset Collection \& Evaluation}
\label{sec:dataset}

We annotated ROC stories~\cite{Mostafazadeh2016ACA} that have corresponding alternative stories in the PASTA dataset~\cite{ghosh2023pasta}.  We selected up to 5 ``story participants,'' defined as volitional entities that were mentioned relatively frequently in the story. This yielded participant-specific story instances for a random 886 actual, unrevised stories from the PASTA dataset. In \cref{app:storyprep} we provide additional details on the processing. %
We annotated all of the participant-specific story instances for these selected stories and all the alternative PASTA stories corresponding to over 50\% of these stories. %
See \cref{tab:document-stats,tab:annotation-details} for high-level dataset and annotation statistics. 


\begin{table}[t]
\centering
\resizebox{.98\columnwidth}{!}{
\small
\begin{tabular}[trim=0 .2in 0 0,  width=.95\linewidth]{p{0.75\linewidth} | p{0.2\linewidth}}
\hline
\# Actual stories annotated & 886  \\
\hline
Avg. volitional participants per story   &   1.12 \\
\hline
Participant specific story instances &   995 \\
\hline
Non-volitional participants (discarded and not included in any of the stats in this table)  & 29 \\
\hline
Actual goal annotation sets   & 2985 \\
\hline
Train/Val/Test splits for Actual sets & 2628/106/219 \\
\hline
\hline
Alternative stories annotated/unannotated   & 951 (934)\\
(corresponding Actual stories)  & 449 (437) \\
\hline
Participant specific alternative instances & 1085 (1027)\\
\hline
Goals applied to alternative stories  &  1527  \\
\hline
\# Alternative goal annotation sets  & 3255\\
\hline
Train/Val/Test splits for alternative sets & 2481/214/512 \\
\hline
\end{tabular}
}
\caption{Document-level data statistics.}
\label{tab:document-stats}
\vspace{-5mm}
\end{table}

\subsection{HIT Design \& Worker Compensation}
\label{sec:HIT_design}
We developed 3 different HITs for our tiered approach to annotation.  We first collect 3 sets of goal annotations for each volitional participant in an actual story (HIT 1), then apply the collected annotations to each corresponding alternative story identifying if the goal is valid using multiple annotator judgements (HIT 2) and for the valid goals obtain new annotation sets suited to the alternative context (HIT3). See \cref{app:HIT_design} and \cref{tab:HITS-annotation} for more details on HIT design. 
Aiming to collect strictly human annotations, we followed the approach used by \newcite{zaidan-callison-burch-2011-crowdsourcing} and converted each participant specific story into an image to disallow easy copying of text into ChatGPT or other such AI tools. Our HIT price was calculated with the aim to pay \$12-\$15 per hour (see \cref{app:compensation} for details).  Our protocol is IRB approved.

\begin{table}[t]
\centering
\resizebox{.98\columnwidth}{!}{
\small
\begin{tabular}[trim=0 .2in 0 0,  width=.98\linewidth]{p{0.60\linewidth} | p{0.19\linewidth} | p{0.20\linewidth}}
\hline
Annotation Type &  Actual & Alternative \\ \hline\hline
Avg. length of goal desc. &   \multicolumn{2}{|c}{5.7 words} \\ \hline
Goals  apply/not to Alter. &\hfil N/A & 89.9/10.1 \%\\
\hline
Goals succeeded/not/unsure/ & 71/16/13 \% & 47/41/7/5 \%\\ 
not-applicable at end of story &  &    \\ \hline\hline
Avg. length of next actions&  6.4 words & 6.8 words\\ \hline
Avg. length of explanations &  7.1 words & 7.6 words \\ \hline
Are Actual story Next Actions transferable/not/unsure to Alternative & \hfil N/A & 54/39/7 \%\\ \hline
Explan. transfer/not to alter..& \hfil N/A & 61/39 \%\\ \hline
goals succeeded/not/unsure & 59/36/5 \%  & 62/33/5 \% \\
after story with Next Action & & \\\hline
Plans Revised/un-revised/cannot-revi  & 65/20/15 \% & 62/26/12 \%\\ \hline
Avg. length of plans & 5.1 words & 5.2 words\\  \hline
Participant satisfied/not/unsure & 70/23/7 \% & 62/32/6 \% \\ 
\hline\end{tabular}
}
\caption{Statistics about goal annotation sets.}
\label{tab:annotation-details}
\vspace{-5mm}
\end{table}


 

\subsection{Dataset Quality and Analysis}
\label{sec:dataeval} 
We assess the quality of our annotations using rational and narrative understanding-based criteria
described below.  Additionally we use the same criteria for evaluating model generations.

Aiming to have completely different stories in our training, dev and test sets, we evaluated all goal annotation sets for all participants for 100 actual and all 209 corresponding alternative stories covering 390 actual and 796 alternative annotation sets ($\approx 20\%$ of the annotations).  We obtained 3 crowd evaluations paying an average of \$.30 per evaluation (see \cref{app:goal_eval} for more details). We obtained an average inter-annotator agreement of 80\% for all annotated features, using a weighted Fleiss's Kappa~\cite{Marasini2016AssessingTI}, indicating very high agreement. 
Our experts' evaluations show high agreement with the crowd, indicating high quality annotations; \cref{tab:evaluation-IAA} shows per-feature scores.

\paragraph{Story Goal Knowledge Evaluation Criteria:}  
\label{sec:goal_eval}
Based on the argument that goals are rational statements that should be logical ~\cite{graesser_how,Setiya2011INTENTIONPA,Grice1971-GRIIAU,Davidson1963-DAVARA-6}, we extend the two qualities required of a narrative to be logical~\cite{graesser_how,pennington} to the \textbf{goal description}: ~\textit{coherence} and \textit{explainability}.   We evaluate the goal description on \textit{coherence} which measures how consistent the described goal is, i.e., the degree to which it makes sense, without any conflicting or made up information.  \textit{explainability} measures whether the participant's purposive actions, even when unsuccessful, are addressed (and can be explained) by the goal description.  Since these qualities inherently require goals to be truthful, faithful and purposive, we explicitly evaluate if the goal description is \textit{truthful} and \textit{faithful} for the story context and if it reflects the participant's \textit{intentionality}.  We also evaluate whether the \textbf{goal achievement} is accurate based on the story information.  

\paragraph{Prospective Knowledge Evaluation Criteria:} 
\label{sec:prospective_eval}
The \textbf{Next Action} and \textbf{Future Plan} are extensions to the story and as such should be consistent with it, using the same entities, time and space constraints as the story.  We use \textit{Cohesion} to measure the consistency with story events and any dissonance with story events lowers this score.  Additionally, the \textbf{Next Action} should be logically consistent with the story without contradicting common sense or story information when using new information and we measure this with \textit{Coherence}.  The \textbf{Future Plan} should also be logically consistent with the participant's goal without contradicting story information and provide an appropriate goal achievement plan which we measure with \textit{Correctness}.  We verify the suitability of \textbf{Explanation} to justify the next action and the correctness of goal achievement direction, participant's satisfaction and plan type annotations.

\subsection{Data Splits}
Our evaluation resulted in quality scores for 13 features for each annotation set. While we use these to ensure high overall quality, given the depth of knowledge we are probing, there can be legitimate nuance and variability in reasonable responses. We therefore further use these scores to create our train/dev/test splits. Our process resulted in evaluation (dev/test) sets where a majority of workers had high agreement on \textit{all} 13 features. We provide extensive details of our scoring process in \cref{app:quality-analysis} and data quality details in \cref{tab:splits-IAA}.  

\section{Models \& Evaluation}
\label{sec:models}

\subsection{Models} We examine a number of well-known pretrained and large language models.  

\paragraph{Instruction Fine-tuned with RLHF Language Models}
Both GPT-3.5 Turbo and GPT-4 models are powered by IntentGPT~\cite{ouyang2022training}, a model architecture that was fine-tuned to generate outputs aligned with human intentions of being helpful, truthful and harmless, when responding to a query.  We want test to what extent this alignment with human intentions extends to goals reasoning.   

\paragraph{Multi-task trained Language Models}
Flan-T5~\cite{chung2022scaling} and T5~\cite{JMLR:v21:20-074} are text-to-text models that can be used for any task that can be converted into textual input and output format. T5 uses transfer learning where the model is pre-trained on a mixture of unsupervised and supervised tasks in a multi-task setting.  Flan-T5 extends T5 pre-training with fine-tuning on 1.8K tasks and incorporates chain-of-thought prompting, both of which improve model performance on human instruction following tasks~\cite{wu-etal-2023-chain}.  Since instruction following hinges on understanding intent we want to test whether this extends to goal reasoning.  We want to explore to what extent Story Cloze (a task based on ROC Stories that is part of the 1.8K tasks) influences goal inferences from our dataset.

\subsection{Evaluation Metrics}
\label{sec:model_eval_metrics}
We use both manual and automated metrics for our text generation tasks (Tasks 1, 3, and 4).  Our automated metrics consist of the classic 
\textbf{ROUGE}~\cite{lin-2004-rouge}, \textbf{METEOR}~\cite{lavie-agarwal-2007-meteor}, corpus and Google's version of sentence \textbf{BLEU}~\cite{BLEU}) and \textbf{BertScore}~\cite{bert-score} for text and \textbf{F1}, \textbf{weighted F1} and \textbf{macro F1} for NLI tasks 2 and 5.  We compute these metrics using the evaluate module in HuggingFace; we compute the score for the generated text using 3 reference labels (obtained from the 3 annotations per story participant) and report results for the test split (and some dev split results in the appendix).   

For \textbf{Human Evaluation} we randomly sample 100 test and 50 dev generations taking one from each story (equal numbers from actual and alternative story contexts).  We evaluate these using the same criteria we used for annotations described in \cref{sec:dataeval} with 3 workers.  For each evaluated feature we report the overall average of Likert scores for all samples and the number of generations where the average score from the workers is $\geq 3$. 

\section{Inferences from Goal Knowledge}
\label{sec:Tasks}

Annotations in our dataset mostly consist of implied information often unstated in the story.   Inferring a participant's goal, future goal achieving actions and plans across a variety of alternative story contexts requires a deeper understanding of text using commonsense and world knowledge.  We present a variety of goal based inference tasks that benchmark GPT-3.5 Turbo, GPT-4 and various sizes of Flan-T5 and T5 using prompting and fine-tuning.  Refer to \cref{tab:task-prompts} for our task prompts. 

We explored in-context learning using 0 to 6 examples (balancing labels in classification tasks)
. Performance improves slightly with more examples but a 3-shot setting allows for an even comparison across models; GPT models produce long multi-sentence generations in a zero-shot setting, unlike the short Flan-T5 generations; token length of a base Flan-T5 and T5 limits us to a max of 3 complete examples without any text cut-off.  T5-11b model generates additional gibberish text along with task related generation making human evaluation cumbersome; we only report automated scores for this model.
We fine-tuned several sizes of Flan-T5 and the T5 base model on our training dataset consisting of both actual and alternative stories.\footnote{Due to memory limitations, we were unable to fine-tune larger models like the 11B Flan-T5-XXL model on a single 80GB A100. Nevertheless, our results demonstrate that fine-tuning smaller models can lead to strong performance.} 

We compare all models and settings using automated evaluation metrics described in \cref{sec:model_eval_metrics} and some models using our human evaluation metrics.  We present human evaluation metrics in the main paper and present automated metrics in the appendix. 
The human evaluated generations across the various models are based on the same stories, but their reference goal annotations may differ (for these we report variance for the averaged scores).
Scores significantly\footnote{Statistical significance was computed using \texttt{mlxtend}} lower than reference scores with a p-value $<.05$ are underlined.

\begin{table}[t]
 \centering
\resizebox{.98\columnwidth}{!}{
 \begin{tabular}{|l|c|c|c|c|c|}
 \hline
 &\multicolumn{5}{|c|}{\textbf{Average Likert Scores (\# evaluations with score $\geq$ 3.0)}}\\
Model Type  &Coherence&Explainable&Faithful&Truthful&Intentional\\ 
\hline

Ref Avg ($\sigma^2 <.01$)  &4.61 (50) &4.54 (49) &4.68 (50) &4.73 (50) &4.39 (47) \\ 
\hline
 fT5b   (3-shot) &\underline{3.91} (42)&\underline{4.12} (44)&\underline{4.32} (45) &\underline{4.22} (45)&\underline{3.74} (39) \\
 fT5xxl (3-shot)&\underline{4.35} (48)&4.50 (49)&4.71 (50)&4.66 (49)&\underline{4.00} (40) \\
 gpt3.5t (3-shot)&\underline{4.39} (48)& 4.44 (48)&4.83 (50)&4.69 (49)&\underline{3.83} (36) \\
 gpt4  (3-shot) &\underline{4.35} (46)&\underline{4.31} (46)&4.77 (49)&4.74 (50)&\underline{3.87} (38)\\
\hline
T5b-ft (0-shot)   &\underline{3.82} (39)&\underline{3.86} (38)&\underline{4.03} (41)&\underline{3.99} (41)&\underline{3.63} (35)\\
fT5b-ft (0-shot) &\underline{3.71} (37)&\underline{3.74} (37)&\underline{3.93} (38)&\underline{3.94} (38)&\underline{3.65} (35)\\
fT5b-ft (3-shot)  &4.41 (46)&4.39 (46)& 4.57 (47) & 4.56 (47)&4.29 (47) \\
\hline
\end{tabular}
}
 
 \caption{\textbf{Task 1} results: human evaluation of model-generated goals (for volitional participants in \textbf{actual stories}). See \cref{sec:model_eval_metrics,sec:goal_eval}  for evaluation details, and \cref{tab:goal-manual-metrics-dev,tab:goal-automated-metrics} (in appendix) for more results.
}
\label{tab:goal-manual-metrics}
\vspace{-5mm}
\end{table}

\subsection{Task 1: Goal Inference }
\label{task1}
In this task, we compare model performance on goal generation for an actual story context.  This is a standard sequence generation task where models generate a goal for a given $(S^a, P_i)$. 

\emph{\textbf{Are models capturing the intent behind participants' actions?}}
Results in \cref{tab:goal-manual-metrics} show larger models are better at generating goals that  explain a participant's actions staying mostly truthful and faithful to the story.  
GPT models tend to summarize the story capturing most details, although, GPT-4 generations are more succinct.  We find that succinct goals (from both GPT-4 and Flan-T5-XXL) are more overarching, capturing intentionality better, but, are likely to be less truthful and faithful to the story due to their lack of expressivity when compared to the more descriptive GPT-3.5 Turbo generations.   Pretrained Flan-T5 generations are generally succinct but the base model performs poorly as the identified goals tend to focus on later story events leading to reactionary goals and do not have a participant-focus leading to wrong goals in multi-participant stories.   Fine-tuning the base T5 and Flan-T5 lead to more participant-focused generations but  still contained reactionary goals. 3-shot prompting helped the fine-tuned model to generate overarching, faithful and truthful goals surpassing the performance of larger models (as seen in model generation examples in \cref{tab:model-goal-generations}).  Both fine-tuning and in-context examples help a model capture the intent behind a participant's actions.  Automated results do not reveal any differences; see \cref{tab:goal-automated-metrics} and \cref{tab:goal-manual-metrics} for  additional models and  metrics. Supplemental examples are in \cref{tab:model-goal-generations}.

\begin{table}[t]
 \centering
\resizebox{.98\columnwidth}{!}{
 \begin{tabular}{|l|c|c|c|}
\hline
Model& Overall  (\textbf{F1})& Full Agr. (\textbf{F1})& Par. Agr. (\textbf{F1})\\
\hline
Baseline  &.31 & .17 &.67 \\
\hline
fT5l (0- \& 3-shot)  &.27&.30&.25\\
fT5xxl (0- \& 3-shot)&.60& .50 &.71\\
gpt3.5t (0-shot)&.49&.39&.72\\
gpt3.5t (3-shot)&.53&.41&.67\\
gpt4  (0- \& 3-shot)&.59 &.48 &.72\\
\hline
fT5b-ft (0- \& 3-shot)&.44&.46&.41\\
fT5l-ft (0- \& 3-shot)& .65 & .71 &.61\\
fT5xl-ft (0- \& 3-shot)& .76 & .84 &.70 \\
\hline
\end{tabular}
}
 \caption{\textbf{Task 2} results: goal transferability (comparing performance in full and partial agreement) to \textbf{alternative stories}. The baseline in the first line only predicts ``not transferable.'' Our fine-tuned models generally outperform larger pretrained LMs. See \cref{tab:goal-applicability-metrics-more} (in appendix) for additional models and detailed F1 scores.
}
\label{tab:goal-applicability-metrics}
\vspace{-5mm}
\end{table}

\subsection{Task 2. Inference Transferability} 
\label{task2}
We compare models at identifying the transferability of goal from an actual to an alternative story (a binary inference where a model identifies if goal $G_{ij}$ obtained from $S^a$ is applicable to $S^{c_k}$). Since identifying `a goal that is not applicable to a story' belongs to negative knowledge that LLMs struggle with~\cite{chen2023say,hossain-etal-2022-analysis}, we evaluate models on identifying the negative label (`a goal not applicable to the alternative story').  


\emph{\textbf{Can models transfer goal inference to alternative stories?}}  Larger models are better than smaller models especially at identifying positive labels in full agreement situations as seen in \cref{tab:goal-applicability-metrics} and \cref{tab:goal-applicability-metrics-more}.   
Lower performance on this task is attributable to two aspects. 
(1) We assigned labels through a majority label assignment; while this works in many cases, some stories \textit{are} nuanced and open to legitimate interpretation. This nuance is reflected in both label types leading to an overall lower model performance. (2) Goals that do not apply to the alternative story generally require additional reasoning (such as consideration of story conditions that rule out goal applicability) making the negative label identification more difficult.
In-context examples lead to an improvement in GPT models but do not change Flan-T5 model predictions.  Comparing performance in full and partial agreement settings we show that fine-tuning on our data boosts model performance 
helping smaller models outperform larger pretrained models. 

\subsection{Task 3: Explainable Next Actions }
\label{task3a}
We consider the task of generating an action most likely to happen after the story and justifying it taking a participant-centric view. This is a standard sequence generation task, where for a story and selected participant ($(S^a, P_i)$ or $(S^{c_k}, P_i)$) a model generates explainable actions. These are actions followed by the explanation with the connecting phrase ``and the reason for this action is.''

\begin{table}[t]
 \centering
\resizebox{.98\columnwidth}{!}{
 \begin{tabular}{|l|c|c|c|c|c|c|}
 \hline
&\multicolumn{6}{|c|}{\textbf{Average Likert Scores (\# evaluations with score $\geq$ 3.0)}}\\
& \multicolumn{3}{|c|}{\textbf{Actual Stories}} & \multicolumn{3}{|c|}{\textbf{Alternative Stories}}\\ 
 Model Type & Coherence & Cohesion & Explain. & Coherence & Cohesion & Explain.\\
\hline
 Ref ($\sigma^2 <.01$) & 4.63 (49)& 4.51 (49) & 4.74 (50) & 4.51 (49) & 4.48 (48) & 4.34 (47)\\
 \hline
 fT5b (3-shot) &\underline{2.38} (8)& \underline{2.14} (8)& \underline{1.68} (3)&\underline{2.14} (10)&\underline{2.08} (7)&\underline{1.86} (4)\\
 fT5xxl (3-shot)&4.48 (49)&4.37 (49)&\underline{3.91} (43)&\underline{3.82} (38)&\underline{3.71} (37)&\underline{3.58} (34)\\
 gpt3.5t (3-shot)&4.76 (50) &4.67 (49) &4.79 (48)&4.53 (48)&4.41 (49)&4.70 (49)\\
 gpt4 (3-shot) &4.77 (50)&4.67 (49)& 4.79 (49)&4.55 (48)&4.48 (48)&4.77 (50)\\
 \hline
T5b-ft (0-shot) &\underline{2.97} (23)&\underline{2.91} (23)&\underline{2.97} (25)&\underline{3.51} (33)&\underline{3.44} (32)&\underline{3.20} (25)\\
fT5b-ft  (0-shot) &\underline{3.53} (32)&\underline{3.44} (31)&\underline{3.42} (32)&\underline{3.67} (38)&\underline{3.65} (36)&\underline{3.30} (30)\\
fT5b-ft  (3-shot) &\underline{2.42} (15) &\underline{2.43} (14)&\underline{2.51} (18)&\underline{2.74} (15)&\underline{2.63} (14)&\underline{2.53} (14)\\
\hline
\end{tabular}
}
 \caption{\textbf{Task 3a} results: human evaluation of model generated Next Actions with Explanations (generations containing both were evaluated). 
 Scores underlined  are significantly lower than reference.  See \cref{sec:model_eval_metrics,sec:goal_eval}  for evaluation details, and \cref{tab:next-actions-manual-metrics-dev,tab:next-action-automated-metrics} (in appendix) for additional models and metrics.%
 }
\label{tab:next-actions-manual-metrics}
\vspace{-5mm}
\end{table}

\emph{\textbf{Can models generate explainable participant-centric actions?}}  Larger models generate more coherent and cohesive next actions; GPT models generate good explanations as seen in \cref{tab:next-actions-manual-metrics} although 3\% are code generations instead of text. 
Pretrained Flan-T5 models struggle to generate explanations with the actions: the base model generates explanations for only 60\% of actions and XXL for 90\%. %

The performance for generating next actions for alternative stories was comparatively lower than for actual stories. %
We noticed that larger models tended to infer next actions for the alternative stories that were not consistent with the alternative story, but rather consistent with the original story. %
We show examples in \cref{tab:model-action-generations}. %
We note that ROCStories are part of the fine-tuning instruction sets used in the  Flan-T5 model pretraining ~\cite{longpre2023flan,wang-etal-2022-super}; it is possible this leads to some degree of memorization and impacts the generalization ability of models. Smaller models fine-tuned on our data do not contain these errors and generate explanations for all actions. 
However, 3-shot prompting of fine-tuned models does not help on this task as it leads to generations that repeat story actions; lower scores reflect this logical inconsistency. 
See \cref{tab:next-actions-manual-metrics-dev,tab:next-action-automated-metrics} for more results.

\label{task3b}
\paragraph{Inference Transferability:} Since next actions and their explanations from $S^a$ are not always logical when applied to $S^{c_k}$, our annotations include updated next actions and explanations. We cast identifying whether the next action is most likely, somewhat likely or unlikely as a 3-way entailment given $S^c_k$ and the next action from $S^a$.   Similarly we cast identifying whether an explanation logically justifies the next action as a binary entailment. 

\begin{table}[t]
 \centering
\resizebox{.98\columnwidth}{!}{
 \begin{tabular}{|l|c|c|c|}
\hline
&\multicolumn{2}{|c|}{\textbf{Task 3b}} & \textbf{Task 3c} \\
&\multicolumn{2}{|c|}{\textbf{Next Action}} & \textbf{Explanation} \\
Model Type & macro F1 & wt. F1& wt. F1\\
\hline
Majority label  &.48&.91&.57\\
\hline
fT5l (0- \& 3-shot)  &.37&.60&.78\\
fT5xxl (0- \& 3-shot)&.45&.67&.79\\
gpt3.5t (0-shot)&.33&.47&.62\\
gpt3.5t (3-shot)&.37&.48&.59\\
gpt4  (0- \& 3-shot)&.47&.72&.84\\
\hline
fT5b-ft (0- \& 3-shot)&.43&.67&.70 \\
fT5l-ft (0- \& 3-shot)&{.49}&.{74}&.75 \\
\hline
\end{tabular}
}
 \caption{\textbf{Task 3b, 3c} results: next actions and explanations transferability. 
 See \cref{tab:next-action-applicability-metrics-more} for more models and detailed F1 scores.
}
\label{tab:next-action-applicability-metrics}
\vspace{-5mm}
\end{table}

\emph{\textbf{Are models able to identify participant-centric next actions and justifications?}} Although larger models are better at generating next actions, they struggle to identify the likelihood of next action classifying most as ``unsure'' (see  \cref{tab:next-action-applicability-metrics} and, in the appendix, \cref{tab:next-action-applicability-metrics-more}). They are better at identifying justifications as with the generations, although, the performance on positive labels is better than the negative labels.  Fine-tuning on our dataset helps smaller models reach or surpass larger models.

\subsection{Task 4. Goal Achievement Plan Inference}
\label{task4}
This task generates a goal achievement plan (for the unachieved goals, roughly a fourth in actual stories and a half in alternate stories) and identifies whether it is based on the participant's original intent or revised based on the story outcome. We found joint generation caused all models to incorrectly identify nearly all plans as revised. This suggests a potential limitation in the models. We therefore examine generating a plan and identifying a plan type separately.  
 
\emph{\textbf{Do models generate reasonable plans?}}  
We show in \cref{tab:future-plan-manual-metrics} that fine-tuning on our data helps smaller models generate plans score on par with GPT plan generations; they are better at identifying plan type but are unable to generate a plan reflecting the correct type. 
GPT models generate generic plans which did not always address the specific situation and they are not of the expected plan type. Fine-tuning generates story specific plans and more of the expected plan type as indicated by the human evaluation Type score. See examples in \cref{tab:model-plan-generations} and additional metrics and models in
\cref{tab:future-plan-automated-metrics}.

\begin{table}[t]
 \centering
\resizebox{.98\columnwidth}{!}{
 \begin{tabular}{|l|c|c|c|c|c|c|c|c|}
 \hline
 & \multicolumn{3}{|c|}{\textbf{Actual Stories}} & \multicolumn{3}{|c|}{\textbf{Alternative Stories}}\\    &Plan&\multicolumn{2}{|c|}{Human Eval.} & Plan&\multicolumn{2}{|c|}{Human Eval.}\\ 
 Model Type&Type F1& Plan &Type&Type F1& Plan &Type\\
 \hline
\hline
Ref ($\sigma^2 <.01$)/Maj    &.77&4.19 (37)&4.19  (37)&.77&4.10 (48)&4.09 (48) \\ 
fT5b (3-shot)&.37 &\underline{3.50} (31)&\underline{2.92} (21)&.48&\underline{3.09} (30)&\underline{3.01} (28) \\
fT5xxl (3-shot)&.44&\underline{3.26} (24)&\underline{2.24} (08)&.38&\underline{3.57} (37)&\underline{2.75} (24)\\
gpt3.5t (3-shot)&.83&3.87 (33)&\underline{2.35} (13)&.72&3.99 (44)&\underline{2.51} (16)\\
gpt4 (3-shot) &.82&4.16 (38)&\underline{2.46} (13)&.79&4.07 (47)&\underline{2.53} (18)\\
\hline
T5b-ft (0-shot) &.78&3.60 (28)&\underline{3.16} (23)&.83&4.10 (45) &\underline{3.13} (24)\\
fT5b-ft (0-shot)&.79&4.07 (37)&\underline{3.15} (17)&.91&3.72 (40)&\underline{3.33} (26)\\ 
fT5b-ft (3-shot)&.79&\underline{2.96} (21)&\underline{2.30} (08)&.91&\underline{2.78} (16)&\underline{2.46} (13)\\ 
\hline
\end{tabular}
}
 \caption{\textbf{Task 4} results: model-generated plans and plan types.  Ref/Maj lists reference human evaluation for the correctness of plan and type (See \cref{sec:model_eval_metrics} \&  \cref{sec:goal_eval}  for details) and majority class F1 for Plan Type. Underlined scores are significantly lower than reference.  For actual stories only 40 plans were evaluated instead of 50. See  \ref{tab:future-plan-automated-metrics} (in appendix) for more models and metrics.
}
\label{tab:future-plan-manual-metrics}
\vspace{-5mm}
\end{table}

\subsection{Task 5. Goal Achievement Inference}
\label{task5}
We examine whether models can identify when a goal is achieved in the story (Task 5a); (possibly) achieved after the story with the next action (Task 5b); and whether the participant is satisfied with the goal achieving (Task 5c). We cast each task as a 3-way entailment, given $S^a$ (or $S^{c_k}$), a next action with explanation and a participant's goal.

\begin{table}[t]
 \centering
\resizebox{.98\columnwidth}{!}{
 \begin{tabular}{|l|c|c|c|c|c|c|}
 \hline
& \multicolumn{2}{|c|}{\textbf{Task 5a}} & \multicolumn{2}{|c|}{\textbf{Task 5b}} &  \multicolumn{2}{|c|}{\textbf{Task 5c}}\\ 
& \multicolumn{2}{|c|}{\textbf{Within Story (mF1)}} & \multicolumn{2}{|c|}{\textbf{After Story (mF1)}} &  \multicolumn{2}{|c|}{\textbf{Satisfaction (mF1)}}\\ 
 \cline{2-7}
& Actual  & Alter. & Actual & Alter. & Actual & Alter. \\ 
\hline
 Maj.   &.38&.22&.28&.27&.28&.26\\ 
 fT5b (3-shot)   &.28&.32&.41&.33&.35&.36\\
 fT5xxl (0-shot)&.47&.55&.59&.29&.54&.46\\
 fT5xxl (3-shot) &.47&.55&.58&.32&.52&.45\\
 gpt3.5t(0-shot) &.41&.51&.60&.27&.37&.33\\
 gpt3.5t(3-shot) &.42&.54&.58&.44&.53&.47\\
 gpt4 (0-shot)   &.45&.66&.43&.49&.50&.44\\
 gpt4 (3-shot)   &.46&.68&.41&.56&.56&.49\\
 \hline
fT5l-ft (0-shot)&.45&.53 &.58&.46&\textbf{.60}&\textbf{.52}\\
fT5xl-ft (0-shot)&.47& .56&.59&.52&\textbf{.65}&\textbf{.75}\\
 \hline 
\end{tabular}
}
 \caption{\textbf{Task 5} results: model identification of  achievement within \& after story and participant satisfaction. See \cref{tab:in-story-metrics,tab:after-story-metrics,tab:after-story-metrics-more} (in appendix) for more results.
 }
\label{tab:goal-achievement-metrics}
\vspace{-6mm}
\end{table}

\emph{\textbf{Are models able to distinguish between implied details across story variations?}}    Identifying goal achievement within a story is limited by a model's ability to identify implied achievement and degree of achievement. As seen in the first 2 columns of \cref{tab:goal-achievement-metrics}, larger models are better, but smaller models improve with fine-tuning.  Achievement is implied in fewer alternative stories than actual stories leading to more negative labels which larger models are better at identifying.  GPT models improve with 3-shot prompting and they are able to identify some of the unsure instances ($<5\%$ are unsure labels) improving performance. Fine-tuned smaller model performance is on par with larger models and are able to identify unsure labels in alternate stories.

The next action after a story helps identify achievable vs. unachievable goals and thus influences a participant's satisfaction.  Larger models are better at identifying after-story achievement and participant's satisfaction for actual stories. 3-shot prompting helps improve smaller models, but the constraints placed by the examples dampen the performance of larger model slightly on identifying achievement but improve the satisfaction identification. All models perform poorly on alternate stories, though some errors that may be attributable to memorization are corrected with 3-shot prompting.  Smaller models fine-tuned on our dataset also identify unsure labels  outperforming larger models. %

\section{Conclusions}
\label{sec:conclusions}
We showed examining goal achievement in alternative stories can lead to a deeper and nuanced understanding of complex events.  Focusing on a specific participant's actions we developed a multi-tiered crowd sourcing process to obtain 6.3k goal annotation sets for 1.3K alternative stories. We captured highly subjective story aspects with our annotations and validated 20\% with high inter-annotator agreement.   We formulated 5 inference tasks and several sub-tasks  to evaluate current SOTA intent-following LLMs.  Our evaluations show that each model differs in specific aspects of goal reasoning providing multiple future avenues to study.  We think modeling advances can achieve a broader and deeper narrative understanding and hope that our work can help further this research.

\section{Limitations}
\label{sec:limit}
We acknowledge our work has the following limitations:
\begin{enumerate}
\item While our annotations are based on well known NLP data sources, our efforts focus on more formal written english. We tried to control for human subjectivity when trying to identify with participants in the stories we annotated through specific instructions. 
\item We use pre-trained large language models in our experiments. These models can echo biases and mis-information either implicitly or explicitly. We do not attempt to control for these in this work.  
\item Model generation and classification abilities can vary as the formality, style, or language change across the crowd written stories we annotated.
\end{enumerate}

\section*{Acknowledgements}
We would like to thank the anonymous reviewers for their comments, questions, and suggestions. %
We would also like to thank Sayontan Ghosh and Niranjan Balasubramian for early discussions and feedback. %
This material is based in part upon work supported by the National Science Foundation under Grant No. IIS-2024878. %
Some experiments were conducted on the UMBC HPCF, supported by the National Science Foundation under Grant No. CNS-1920079. %
This material is also based on research that is in part supported by the Army Research Laboratory, Grant No. W911NF2120076, and by the Air Force Research Laboratory (AFRL), DARPA, for the KAIROS program under agreement number FA8750-19-2-1003. The U.S. Government is authorized to reproduce and distribute reprints for Governmental purposes notwithstanding any copyright notation thereon. The views and conclusions contained herein are those of the authors and should not be interpreted as necessarily representing the official policies or endorsements, either express or implied, of the Air Force Research Laboratory (AFRL), DARPA, or the U.S. Government. %

\bibliography{anthology,custom}

\begin{thebibliography}{62}
\expandafter\ifx\csname natexlab\endcsname\relax\def\natexlab#1{#1}\fi

\bibitem[{Ajzen and Kruglanski(2019)}]{reasoned_action}
Icek Ajzen and Arie Kruglanski. 2019.
\newblock \href {https://doi.org/10.1037/rev0000155} {Reasoned action in the
  service of goal pursuit}.
\newblock \emph{Psychological Review}, 126.

\bibitem[{Ammanabrolu et~al.(2021)Ammanabrolu, Urbanek, Li, Szlam,
  Rockt{\"a}schel, and Weston}]{ammanabrolu-etal-2021-motivate}
Prithviraj Ammanabrolu, Jack Urbanek, Margaret Li, Arthur Szlam, Tim
  Rockt{\"a}schel, and Jason Weston. 2021.
\newblock How to motivate your dragon: Teaching goal-driven agents to speak and
  act in fantasy worlds.
\newblock In \emph{Proceedings of the 2021 Conference of the North American
  Chapter of the Association for Computational Linguistics: Human Language
  Technologies}.

\bibitem[{Austin and Vancouver(1996)}]{Goal_Constructs_in_psychology}
J.~T. Austin and J.~B. Vancouver. 1996.
\newblock Goal constructs in psychology: Structure, process, and content.
\newblock \emph{Psychological Bulletin}, 120(3):338--375.

\bibitem[{Bellos et~al.(2024)Bellos, Li, Liu, and
  Corso}]{bellos-etal-2024-large}
Filippos Bellos, Yayuan Li, Wuao Liu, and Jason Corso. 2024.
\newblock \href {https://aclanthology.org/2024.scalellm-1.3} {Can large
  language models reason about goal-oriented tasks?}
\newblock In \emph{Proceedings of the First edition of the Workshop on the
  Scaling Behavior of Large Language Models (SCALE-LLM 2024)}, pages 24--34,
  St. Julian{'}s, Malta. Association for Computational Linguistics.

\bibitem[{Binswanger(1991)}]{BINSWANGER1991154}
Harry Binswanger. 1991.
\newblock \href {https://doi.org/https://doi.org/10.1016/0749-5978(91)90019-P}
  {Volition as cognitive self-regulation}.
\newblock \emph{Organizational Behavior and Human Decision Processes},
  50(2):154--178.
\newblock Theories of Cognitive Self-Regulation.

\bibitem[{Brown et~al.(2020)Brown, Mann, Ryder, Subbiah, Kaplan, Dhariwal,
  Neelakantan, Shyam, Sastry, Askell, Agarwal, Herbert-Voss, Krueger, Henighan,
  Child, Ramesh, Ziegler, Wu, Winter, Hesse, Chen, Sigler, Litwin, Gray, Chess,
  Clark, Berner, McCandlish, Radford, Sutskever, and Amodei}]{brown}
Tom~B. Brown, Benjamin Mann, Nick Ryder, Melanie Subbiah, Jared Kaplan,
  Prafulla Dhariwal, Arvind Neelakantan, Pranav Shyam, Girish Sastry, Amanda
  Askell, Sandhini Agarwal, Ariel Herbert-Voss, Gretchen Krueger, Tom Henighan,
  Rewon Child, Aditya Ramesh, Daniel~M. Ziegler, Jeffrey Wu, Clemens Winter,
  Christopher Hesse, Mark Chen, Eric Sigler, Mateusz Litwin, Scott Gray,
  Benjamin Chess, Jack Clark, Christopher Berner, Sam McCandlish, Alec Radford,
  Ilya Sutskever, and Dario Amodei. 2020.
\newblock Language models are few-shot learners.
\newblock In \emph{Proceedings of the 34th International Conference on Neural
  Information Processing Systems}, NIPS'20, Red Hook, NY, USA. Curran
  Associates Inc.

\bibitem[{Callison-Burch et~al.(2022)Callison-Burch, Tomar, Martin, Ippolito,
  Bailis, and Reitter}]{callison-burch-etal-2022-dungeons}
Chris Callison-Burch, Gaurav~Singh Tomar, Lara Martin, Daphne Ippolito, Suma
  Bailis, and David Reitter. 2022.
\newblock Dungeons and dragons as a dialog challenge for artificial
  intelligence.
\newblock In \emph{Proceedings of the 2022 Conference on Empirical Methods in
  Natural Language Processing}.

\bibitem[{Cao et~al.(2022)Cao, Lu, DeFazio, and Zhang}]{cao-etal-2022-goal}
Yan Cao, Keting Lu, David DeFazio, and Shiqi Zhang. 2022.
\newblock \href {https://doi.org/10.18653/v1/2022.findings-emnlp.327}
  {Goal-oriented vision-and-dialog navigation via reinforcement learning}.
\newblock In \emph{Findings of the Association for Computational Linguistics:
  EMNLP 2022}, pages 4473--4482, Abu Dhabi, United Arab Emirates. Association
  for Computational Linguistics.

\bibitem[{Carpendale and Lewis(2015)}]{carpendale_2015}
Jeremy I.~M. Carpendale and Charlie Lewis. 2015.
\newblock \href
  {https://doi.org/https://doi.org/10.1002/9781118963418.childpsy210}
  {\emph{The Development of Social Understanding}}, chapter~10. John Wiley and
  Sons, Ltd.

\bibitem[{Chen et~al.(2023)Chen, Shi, Fu, Cheng, Li, and Xiao}]{chen2023say}
Jiangjie Chen, Wei Shi, Ziquan Fu, Sijie Cheng, Lei Li, and Yanghua Xiao. 2023.
\newblock \href {http://arxiv.org/abs/2305.05976} {Say what you mean! large
  language models speak too positively about negative commonsense knowledge}.

\bibitem[{Chung et~al.(2022)Chung, Hou, Longpre, Zoph, Tay, Fedus, Li, Wang,
  Dehghani, Brahma, Webson, Gu, Dai, Suzgun, Chen, Chowdhery, Castro-Ros,
  Pellat, Robinson, Valter, Narang, Mishra, Yu, Zhao, Huang, Dai, Yu, Petrov,
  Chi, Dean, Devlin, Roberts, Zhou, Le, and Wei}]{chung2022scaling}
Hyung~Won Chung, Le~Hou, Shayne Longpre, Barret Zoph, Yi~Tay, William Fedus,
  Yunxuan Li, Xuezhi Wang, Mostafa Dehghani, Siddhartha Brahma, Albert Webson,
  Shixiang~Shane Gu, Zhuyun Dai, Mirac Suzgun, Xinyun Chen, Aakanksha
  Chowdhery, Alex Castro-Ros, Marie Pellat, Kevin Robinson, Dasha Valter,
  Sharan Narang, Gaurav Mishra, Adams Yu, Vincent Zhao, Yanping Huang, Andrew
  Dai, Hongkun Yu, Slav Petrov, Ed~H. Chi, Jeff Dean, Jacob Devlin, Adam
  Roberts, Denny Zhou, Quoc~V. Le, and Jason Wei. 2022.
\newblock \href {http://arxiv.org/abs/2210.11416} {Scaling
  instruction-finetuned language models}.

\bibitem[{Davidson(1963)}]{Davidson1963-DAVARA-6}
Donald Davidson. 1963.
\newblock \href {https://doi.org/10.2307/2023177} {Actions, reasons, and
  causes}.
\newblock \emph{Journal of Philosophy}, 60(23):685.

\bibitem[{Davis and Marcus(2015)}]{Davis2015CommonsenseRA}
Ernest Davis and Gary~F. Marcus. 2015.
\newblock Commonsense reasoning and commonsense knowledge in artificial
  intelligence.
\newblock \emph{Communications of the ACM}, 58:92 -- 103.

\bibitem[{Foss and Bower(1986)}]{foss_and_bower_1986}
Carolyn~L Foss and Gordon~H Bower. 1986.
\newblock Understanding actions in relation to goals.
\newblock \emph{Advances in cognitive science}, 1:94--124.

\bibitem[{Geva et~al.(2021)Geva, Khashabi, Segal, Khot, Roth, and
  Berant}]{10.1162/tacl_a_00370}
Mor Geva, Daniel Khashabi, Elad Segal, Tushar Khot, Dan Roth, and Jonathan
  Berant. 2021.
\newblock \href {https://doi.org/10.1162/tacl_a_00370} {{Did Aristotle Use a
  Laptop? A Question Answering Benchmark with Implicit Reasoning Strategies}}.
\newblock \emph{Transactions of the Association for Computational Linguistics},
  9:346--361.

\bibitem[{Ghosh et~al.(2023)Ghosh, Koupaee, Chen, Ferraro, Chambers, and
  Balasubramanian}]{ghosh2023pasta}
Sayontan Ghosh, Mahnaz Koupaee, Isabella Chen, Francis Ferraro, Nathanael
  Chambers, and Niranjan Balasubramanian. 2023.
\newblock \href {http://arxiv.org/abs/2208.00329} {Pasta: A dataset for
  modeling participant states in narratives}.

\bibitem[{Gollwitzer(1993)}]{gollwitzer_intention}
Peter~M. Gollwitzer. 1993.
\newblock \href {https://doi.org/10.1080/14792779343000059} {Goal achievement:
  The role of intentions}.
\newblock \emph{European Review of Social Psychology}, 4(1):141--185.

\bibitem[{Gollwitzer and Bargh(1996)}]{gollwitzer_and_bargh}
{Peter M} Gollwitzer and {John A} Bargh, editors. 1996.
\newblock \emph{The psychology of action: Linking cognition and motivation to
  behavior}.
\newblock Guilford Press.
\newblock Includes bibliographical references and index.

\bibitem[{Graesser et~al.(1994)Graesser, Singer, and Trabasso}]{graesser}
Arthur Graesser, Murray Singer, and Tom Trabasso. 1994.
\newblock \href {https://doi.org/10.1037/0033-295X.101.3.371} {Constructing
  inferences during narrative text comprehension}.
\newblock \emph{Psychological review}, 101:371--95.

\bibitem[{Graesser et~al.(2020)Graesser, Olde, and Klettke}]{graesser_how}
Arthur~C. Graesser, Brent Olde, and Bianca Klettke. 2020.
\newblock \emph{How does the mind construct and represent stories?}
\newblock Lawrence Erlbaum Associates Publishers.

\bibitem[{Grice(1971)}]{Grice1971-GRIIAU}
H.~P. Grice. 1971.
\newblock Intention and uncertainty.
\newblock \emph{Proceedings of the British Academy}, 57:263--279.

\bibitem[{Harmon(2005)}]{harmon}
Mary~E. Harmon. 2005.
\newblock \emph{Factors affecting the activation of predictive inferences}.
\newblock Univeristy of New Hampshire.

\bibitem[{Hossain et~al.(2022)Hossain, Chinnappa, and
  Blanco}]{hossain-etal-2022-analysis}
Md~Mosharaf Hossain, Dhivya Chinnappa, and Eduardo Blanco. 2022.
\newblock \href {https://doi.org/10.18653/v1/2022.acl-short.81} {An analysis of
  negation in natural language understanding corpora}.
\newblock In \emph{Proceedings of the 60th Annual Meeting of the Association
  for Computational Linguistics (Volume 2: Short Papers)}, pages 716--723,
  Dublin, Ireland. Association for Computational Linguistics.

\bibitem[{Hovy and Yang(2021)}]{hovy-yang-2021-importance}
Dirk Hovy and Diyi Yang. 2021.
\newblock \href {https://doi.org/10.18653/v1/2021.naacl-main.49} {The
  importance of modeling social factors of language: Theory and practice}.
\newblock In \emph{Proceedings of the 2021 Conference of the North American
  Chapter of the Association for Computational Linguistics: Human Language
  Technologies}, pages 588--602, Online. Association for Computational
  Linguistics.

\bibitem[{Jiang and Riloff(2018)}]{jiang-riloff-2018-learning}
Tianyu Jiang and Ellen Riloff. 2018.
\newblock \href {https://doi.org/10.18653/v1/P18-1120} {Learning prototypical
  goal activities for locations}.
\newblock In \emph{Proceedings of the 56th Annual Meeting of the Association
  for Computational Linguistics (Volume 1: Long Papers)}, pages 1297--1307,
  Melbourne, Australia. Association for Computational Linguistics.

\bibitem[{Jiang et~al.(2023)Jiang, Ilievski, and Ma}]{jiang}
Yifan Jiang, Filip Ilievski, and Kaixin Ma. 2023.
\newblock \href {http://arxiv.org/abs/2304.13867} {Transferring procedural
  knowledge across commonsense tasks}.

\bibitem[{Joshi et~al.(2019)Joshi, Chen, Liu, Weld, Zettlemoyer, and
  Levy}]{mandar_spanbert}
Mandar Joshi, Danqi Chen, Yinhan Liu, Daniel~S. Weld, Luke Zettlemoyer, and
  Omer Levy. 2019.
\newblock \href {http://arxiv.org/abs/1907.10529} {Spanbert: Improving
  pre-training by representing and predicting spans}.
\newblock \emph{CoRR}, abs/1907.10529.

\bibitem[{Keefe and McDaniel(1993)}]{KEEFE1993446}
Dennis~E. Keefe and Mark~A. McDaniel. 1993.
\newblock \href {https://doi.org/https://doi.org/10.1006/jmla.1993.1024} {The
  time course and durability of predictive inferences}.
\newblock \emph{Journal of Memory and Language}, 32(4):446--463.

\bibitem[{Lal et~al.(2021)Lal, Chambers, Mooney, and
  Balasubramanian}]{lal-etal-2021-tellmewhy}
Yash~Kumar Lal, Nathanael Chambers, Raymond Mooney, and Niranjan
  Balasubramanian. 2021.
\newblock \href {https://doi.org/10.18653/v1/2021.findings-acl.53}
  {{T}ell{M}e{W}hy: A dataset for answering why-questions in narratives}.
\newblock In \emph{Findings of the Association for Computational Linguistics:
  ACL-IJCNLP 2021}, pages 596--610, Online. Association for Computational
  Linguistics.

\bibitem[{Lal et~al.(2022)Lal, Tandon, Aggarwal, Liu, Chambers, Mooney, and
  Balasubramanian}]{lal-etal-2022-using}
Yash~Kumar Lal, Niket Tandon, Tanvi Aggarwal, Horace Liu, Nathanael Chambers,
  Raymond Mooney, and Niranjan Balasubramanian. 2022.
\newblock Using commonsense knowledge to answer why questions.
\newblock In \emph{Proceedings of the 2022 Conference on Empirical Methods in
  Natural Language Processing}, Online and Abu Dhabi, United Arab Emirates.
  Association for Computational Linguistics.

\bibitem[{Lavie and Agarwal(2007)}]{lavie-agarwal-2007-meteor}
Alon Lavie and Abhaya Agarwal. 2007.
\newblock \href {https://aclanthology.org/W07-0734} {{METEOR}: An automatic
  metric for {MT} evaluation with high levels of correlation with human
  judgments}.
\newblock In \emph{Proceedings of the Second Workshop on Statistical Machine
  Translation}, pages 228--231, Prague, Czech Republic. Association for
  Computational Linguistics.

\bibitem[{Lin(2004)}]{lin-2004-rouge}
Chin-Yew Lin. 2004.
\newblock \href {https://aclanthology.org/W04-1013} {{ROUGE}: A package for
  automatic evaluation of summaries}.
\newblock In \emph{Text Summarization Branches Out}, pages 74--81, Barcelona,
  Spain. Association for Computational Linguistics.

\bibitem[{Locke and Latham(2002)}]{locke_2002}
Edwin Locke and Gary Latham. 2002.
\newblock \href {https://doi.org/10.1037/0003-066X.57.9.705} {Building a
  practically useful theory of goal setting and task motivation: A 35-year
  odyssey}.
\newblock \emph{American Psychologist}, 57(9):705--717.

\bibitem[{Locke et~al.(1981)Locke, Shaw, Saari, and Latham}]{locke_1981}
Edwin Locke, Karyll Shaw, Lise Saari, and Gary Latham. 1981.
\newblock \href {https://doi.org/10.1037/0033-2909.90.1.125} {Goal setting and
  task performance: 1969–1980}.
\newblock \emph{Psychological Bulletin}, 90:125--152.

\bibitem[{Longpre et~al.(2023)Longpre, Hou, Vu, Webson, Chung, Tay, Zhou, Le,
  Zoph, Wei et~al.}]{longpre2023flan}
Shayne Longpre, Le~Hou, Tu~Vu, Albert Webson, Hyung~Won Chung, Yi~Tay, Denny
  Zhou, Quoc~V Le, Barret Zoph, Jason Wei, et~al. 2023.
\newblock The flan collection: Designing data and methods for effective
  instruction tuning.
\newblock \emph{arXiv preprint arXiv:2301.13688}.

\bibitem[{Loshchilov and Hutter(2017)}]{DBLP:journals/corr/abs-1711-05101}
Ilya Loshchilov and Frank Hutter. 2017.
\newblock \href {http://arxiv.org/abs/1711.05101} {Fixing weight decay
  regularization in adam}.
\newblock \emph{CoRR}, abs/1711.05101.

\bibitem[{Lyu et~al.(2021)Lyu, Zhang, and Callison-Burch}]{lyu-etal-2021-goal}
Qing Lyu, Li~Zhang, and Chris Callison-Burch. 2021.
\newblock \href {https://aclanthology.org/2021.inlg-1.19} {Goal-oriented script
  construction}.
\newblock In \emph{Proceedings of the 14th International Conference on Natural
  Language Generation}, pages 184--200, Aberdeen, Scotland, UK. Association for
  Computational Linguistics.

\bibitem[{Marasini et~al.(2016)Marasini, Quatto, and
  Ripamonti}]{Marasini2016AssessingTI}
Donata Marasini, Piero Quatto, and Enrico Ripamonti. 2016.
\newblock Assessing the inter-rater agreement for ordinal data through weighted
  indexes.
\newblock \emph{Statistical Methods in Medical Research}, 25:2611 -- 2633.

\bibitem[{Matherly(2018)}]{mturk_article}
Ted Matherly. 2018.
\newblock \href {https://doi.org/10.1108/EJM-07-2017-0491} {A panel for lemons?
  positivity bias, reputation systems and data quality on mturk}.
\newblock \emph{European Journal of Marketing}, 53.

\bibitem[{Mostafazadeh et~al.(2016)Mostafazadeh, Chambers, He, Parikh, Batra,
  Vanderwende, Kohli, and Allen}]{Mostafazadeh2016ACA}
N.~Mostafazadeh, Nathanael Chambers, Xiaodong He, Devi Parikh, Dhruv Batra,
  Lucy Vanderwende, Pushmeet Kohli, and James~F. Allen. 2016.
\newblock A corpus and cloze evaluation for deeper understanding of commonsense
  stories.
\newblock In \emph{NAACL}.

\bibitem[{Niven and Kao(2019)}]{niven-kao-2019-probing}
Timothy Niven and Hung-Yu Kao. 2019.
\newblock \href {https://doi.org/10.18653/v1/P19-1459} {Probing neural network
  comprehension of natural language arguments}.
\newblock In \emph{Proceedings of the 57th Annual Meeting of the Association
  for Computational Linguistics}, pages 4658--4664, Florence, Italy.
  Association for Computational Linguistics.

\bibitem[{Ouyang et~al.(2022)Ouyang, Wu, Jiang, Almeida, Wainwright, Mishkin,
  Zhang, Agarwal, Slama, Ray, Schulman, Hilton, Kelton, Miller, Simens, Askell,
  Welinder, Christiano, Leike, and Lowe}]{ouyang2022training}
Long Ouyang, Jeff Wu, Xu~Jiang, Diogo Almeida, Carroll~L. Wainwright, Pamela
  Mishkin, Chong Zhang, Sandhini Agarwal, Katarina Slama, Alex Ray, John
  Schulman, Jacob Hilton, Fraser Kelton, Luke Miller, Maddie Simens, Amanda
  Askell, Peter Welinder, Paul Christiano, Jan Leike, and Ryan Lowe. 2022.
\newblock \href {http://arxiv.org/abs/2203.02155} {Training language models to
  follow instructions with human feedback}.

\bibitem[{Papineni et~al.(2002)Papineni, Roukos, Ward, and Zhu}]{BLEU}
Kishore Papineni, Salim Roukos, Todd Ward, and Wei-Jing Zhu. 2002.
\newblock \href {https://doi.org/10.3115/1073083.1073135} {Bleu: A method for
  automatic evaluation of machine translation}.
\newblock In \emph{Proceedings of the 40th Annual Meeting on Association for
  Computational Linguistics}, ACL '02, page 311–318, USA. Association for
  Computational Linguistics.

\bibitem[{Pennington and Hastie(1992)}]{pennington}
Nancy Pennington and Reid Hastie. 1992.
\newblock \href {https://doi.org/10.1037/0022-3514.62.2.189} {Explaining the
  evidence: Tests of the story model for juror decision making}.
\newblock \emph{Journal of Personality and Social Psychology}, 62:189--206.

\bibitem[{Qin et~al.(2019)Qin, Bosselut, Holtzman, Bhagavatula, Clark, and
  Choi}]{qin-etal-2019-counterfactual}
Lianhui Qin, Antoine Bosselut, Ari Holtzman, Chandra Bhagavatula, Elizabeth
  Clark, and Yejin Choi. 2019.
\newblock \href {https://doi.org/10.18653/v1/D19-1509} {Counterfactual story
  reasoning and generation}.
\newblock In \emph{Proceedings of the 2019 Conference on Empirical Methods in
  Natural Language Processing and the 9th International Joint Conference on
  Natural Language Processing (EMNLP-IJCNLP)}, pages 5043--5053, Hong Kong,
  China. Association for Computational Linguistics.

\bibitem[{Raffel et~al.(2020)Raffel, Shazeer, Roberts, Lee, Narang, Matena,
  Zhou, Li, and Liu}]{JMLR:v21:20-074}
Colin Raffel, Noam Shazeer, Adam Roberts, Katherine Lee, Sharan Narang, Michael
  Matena, Yanqi Zhou, Wei Li, and Peter~J. Liu. 2020.
\newblock \href {http://jmlr.org/papers/v21/20-074.html} {Exploring the limits
  of transfer learning with a unified text-to-text transformer}.
\newblock \emph{Journal of Machine Learning Research}, 21(140):1--67.

\bibitem[{Rahimtoroghi et~al.(2017)Rahimtoroghi, Wu, Wang, Anand, and
  Walker}]{rahimtoroghi-etal-2017-modelling}
Elahe Rahimtoroghi, Jiaqi Wu, Ruimin Wang, Pranav Anand, and Marilyn Walker.
  2017.
\newblock \href {https://doi.org/10.18653/v1/W17-5543} {Modelling protagonist
  goals and desires in first-person narrative}.
\newblock In \emph{Proceedings of the 18th Annual {SIG}dial Meeting on
  Discourse and Dialogue}, pages 360--369, Saarbr{\"u}cken, Germany.
  Association for Computational Linguistics.

\bibitem[{Setiya(2011)}]{Setiya2011INTENTIONPA}
Kieran Setiya. 2011.
\newblock Intention, plans, and ethical rationalism.
\newblock \emph{Rational and Social Agency: The Philosophy of Michael Bratman},
  pages 56--82.

\bibitem[{Stiennon et~al.(2020)Stiennon, Ouyang, Wu, Ziegler, Lowe, Voss,
  Radford, Amodei, and Christiano}]{stiennon}
Nisan Stiennon, Long Ouyang, Jeff Wu, Daniel~M. Ziegler, Ryan Lowe, Chelsea
  Voss, Alec Radford, Dario Amodei, and Paul~F. Christiano. 2020.
\newblock \href {http://arxiv.org/abs/2009.01325} {Learning to summarize from
  human feedback}.
\newblock \emph{CoRR}, abs/2009.01325.

\bibitem[{Storks et~al.(2021)Storks, Gao, Zhang, and
  Chai}]{storks-etal-2021-tiered-reasoning}
Shane Storks, Qiaozi Gao, Yichi Zhang, and Joyce Chai. 2021.
\newblock \href {https://doi.org/10.18653/v1/2021.findings-emnlp.422} {Tiered
  reasoning for intuitive physics: Toward verifiable commonsense language
  understanding}.
\newblock In \emph{Findings of the Association for Computational Linguistics:
  EMNLP 2021}, pages 4902--4918, Punta Cana, Dominican Republic. Association
  for Computational Linguistics.

\bibitem[{Trabasso and Suh(1993)}]{trabasso}
Tom Trabasso and Soyoung Suh. 1993.
\newblock \href {https://doi.org/10.1080/01638539309544827} {Understanding
  text: Achieving explanatory coherence through on‐line inferences and mental
  operations in working memory}.
\newblock \emph{Discourse Processes}, 16(1-2):3--34.

\bibitem[{Vallurupalli et~al.(2022)Vallurupalli, Ghosh, Erk, Balasubramanian,
  and Ferraro}]{poque}
Sai Vallurupalli, Sayontan Ghosh, Katrin Erk, Niranjan Balasubramanian, and
  Francis Ferraro. 2022.
\newblock \href {https://doi.org/10.48550/ARXIV.2212.02629} {Poque: Asking
  participant-specific outcome questions for a deeper understanding of complex
  events}.

\bibitem[{Wang et~al.(2022)Wang, Mishra, Alipoormolabashi, Kordi, Mirzaei,
  Naik, Ashok, Dhanasekaran, Arunkumar, Stap, Pathak, Karamanolakis, Lai,
  Purohit, Mondal, Anderson, Kuznia, Doshi, Pal, Patel, Moradshahi, Parmar,
  Purohit, Varshney, Kaza, Verma, Puri, Karia, Doshi, Sampat, Mishra, Reddy~A,
  Patro, Dixit, and Shen}]{wang-etal-2022-super}
Yizhong Wang, Swaroop Mishra, Pegah Alipoormolabashi, Yeganeh Kordi, Amirreza
  Mirzaei, Atharva Naik, Arjun Ashok, Arut~Selvan Dhanasekaran, Anjana
  Arunkumar, David Stap, Eshaan Pathak, Giannis Karamanolakis, Haizhi Lai,
  Ishan Purohit, Ishani Mondal, Jacob Anderson, Kirby Kuznia, Krima Doshi,
  Kuntal~Kumar Pal, Maitreya Patel, Mehrad Moradshahi, Mihir Parmar, Mirali
  Purohit, Neeraj Varshney, Phani~Rohitha Kaza, Pulkit Verma, Ravsehaj~Singh
  Puri, Rushang Karia, Savan Doshi, Shailaja~Keyur Sampat, Siddhartha Mishra,
  Sujan Reddy~A, Sumanta Patro, Tanay Dixit, and Xudong Shen. 2022.
\newblock \href {https://doi.org/10.18653/v1/2022.emnlp-main.340}
  {Super-{N}atural{I}nstructions: Generalization via declarative instructions
  on 1600+ {NLP} tasks}.
\newblock In \emph{Proceedings of the 2022 Conference on Empirical Methods in
  Natural Language Processing}, pages 5085--5109, Abu Dhabi, United Arab
  Emirates. Association for Computational Linguistics.

\bibitem[{Wei et~al.(2022)Wei, Bosma, Zhao, Guu, Yu, Lester, Du, Dai, and
  Le}]{wei2022finetuned}
Jason Wei, Maarten Bosma, Vincent Zhao, Kelvin Guu, Adams~Wei Yu, Brian Lester,
  Nan Du, Andrew~M. Dai, and Quoc~V Le. 2022.
\newblock \href {https://openreview.net/forum?id=gEZrGCozdqR} {Finetuned
  language models are zero-shot learners}.
\newblock In \emph{International Conference on Learning Representations}.

\bibitem[{Wu et~al.(2023)Wu, Zhang, and Huang}]{wu-etal-2023-chain}
Dingjun Wu, Jing Zhang, and Xinmei Huang. 2023.
\newblock \href {https://doi.org/10.18653/v1/2023.findings-acl.408} {Chain of
  thought prompting elicits knowledge augmentation}.
\newblock In \emph{Findings of the Association for Computational Linguistics:
  ACL 2023}, pages 6519--6534, Toronto, Canada. Association for Computational
  Linguistics.

\bibitem[{Zacks(2020)}]{zacks_event}
Jeffrey~M. Zacks. 2020.
\newblock \href {https://doi.org/10.1146/annurev-psych-010419-051101} {Event
  perception and memory}.
\newblock \emph{Annual Review of Psychology}, 71(1):165--191.
\newblock PMID: 31905113.

\bibitem[{Zaidan and
  Callison-Burch(2011)}]{zaidan-callison-burch-2011-crowdsourcing}
Omar~F. Zaidan and Chris Callison-Burch. 2011.
\newblock \href {https://aclanthology.org/P11-1122} {Crowdsourcing translation:
  Professional quality from non-professionals}.
\newblock In \emph{Proceedings of the 49th Annual Meeting of the Association
  for Computational Linguistics: Human Language Technologies}, pages
  1220--1229, Portland, Oregon, USA. Association for Computational Linguistics.

\bibitem[{Zellers et~al.(2018)Zellers, Bisk, Schwartz, and Choi}]{zellers2018}
Rowan Zellers, Yonatan Bisk, Roy Schwartz, and Yejin Choi. 2018.
\newblock \href {https://doi.org/10.18653/v1/D18-1009} {Swag: A large-scale
  adversarial dataset for grounded commonsense inference}.
\newblock In \emph{EMNLP}, pages 93--104.

\bibitem[{Zellers et~al.(2019)Zellers, Holtzman, Bisk, Farhadi, and
  Choi}]{Zellers2019HellaSwagCA}
Rowan Zellers, Ari Holtzman, Yonatan Bisk, Ali Farhadi, and Yejin Choi. 2019.
\newblock \href {https://api.semanticscholar.org/CorpusID:159041722}
  {Hellaswag: Can a machine really finish your sentence?}
\newblock In \emph{Annual Meeting of the Association for Computational
  Linguistics}.

\bibitem[{Zhang et~al.(2020{\natexlab{a}})Zhang, Lyu, and
  Callison-Burch}]{zhang-etal-2020-reasoning}
Li~Zhang, Qing Lyu, and Chris Callison-Burch. 2020{\natexlab{a}}.
\newblock \href {https://doi.org/10.18653/v1/2020.emnlp-main.374} {Reasoning
  about goals, steps, and temporal ordering with {W}iki{H}ow}.
\newblock In \emph{Proceedings of the 2020 Conference on Empirical Methods in
  Natural Language Processing (EMNLP)}, pages 4630--4639, Online. Association
  for Computational Linguistics.

\bibitem[{Zhang et~al.(2023)Zhang, Naradowsky, and Miyao}]{zhang2023ask}
Qiang Zhang, Jason Naradowsky, and Yusuke Miyao. 2023.
\newblock \href {http://arxiv.org/abs/2305.17878} {Ask an expert: Leveraging
  language models to improve strategic reasoning in goal-oriented dialogue
  models}.

\bibitem[{Zhang et~al.(2020{\natexlab{b}})Zhang, Kishore, Wu, Weinberger, and
  Artzi}]{bert-score}
Tianyi Zhang, Varsha Kishore, Felix Wu, Kilian~Q. Weinberger, and Yoav Artzi.
  2020{\natexlab{b}}.
\newblock \href {https://openreview.net/forum?id=SkeHuCVFDr} {Bertscore:
  Evaluating text generation with bert}.
\newblock In \emph{International Conference on Learning Representations}.

\end{thebibliography}
\bibliographystyle{acl_natbib}

\newpage
\clearpage
\appendix
\section{Appendix}
\label{sec:appendix}

No AI assistants were used for writing either this paper or code for producing the work in this paper.  All writing is original and produced by the authors.

\subsection{Story Preparation}
\label{app:storyprep}
This section provides additional details for the annotation protocol described in \cref{sec:dataset}. %

For stories in the ROC story collection, we  identified entities using NER\footnote{using Spacy https://spacy.io/} and selected the types that are capable of employing volition to influence or achieve a desired state.\footnote{volition capable types include EVENT, LAW, PERSON, NORP \& ORG} In each story, we identified volitional entities using dependency parsing~\footnote{volition-capable entities that are subjects of predicates} and identified any co-referring entity mentions using an automatic entity coreference system\footnote{using SpanBERT~\cite{mandar_spanbert}}.  We selected  up to 5 volitional entities (aka. story participants) with the most mentions in a story and created participant-specific story instances. 

As a first step in the annotation process, 3 crowd workers verified that the selected participants are volitional entities.  When a worker identified a participant as non-volitional, we verified and discarded the participant-specific story (Total discarded :29). For the remaining 995 participant specific stories, we obtained 3 goal annotations and for a random 50\% of these stories we annotated all the corresponding counterfactual (in our terminology, ``alternative'') stories in PASTA.  We obtained a total of 2985 actual and 3240 alternative goal annotations using a chain of Human Intelligence Tasks (HITs) to identify a participant's goal in story and follow its achievement-arc in both the actual and alternative stories (for the annotated alternative stories).  See \cref{sec:HIT_design} for HIT design and compensation.

\begin{table*}[]
\centering
\resizebox{.98\textwidth}{!}{
\small
\begin{tabular}[trim=0 .2in 0 0,  width=.95\linewidth]{p{0.19\linewidth} | p{0.60\linewidth} |p{0.19\linewidth} 
}
\hline\hline
Annotation & Description of the Annotated Knowledge & Example from \cref{fig:fig3}\\
\hline 
\multicolumn{2}{c}{\textbf{Goal Knowledge Annotation with in Actual stories with HIT 1}} \\
\hline 
Volitional Participant $P_i$ &  This annotation provides a binary decision of whether the participant aims to achieve a goal in the actual story $S^a$.  Further annotation of the story continues only if the participant is volitional.  & Yes and the participant is ``her kids''. \\
\hline 
Goal Description $G_{ij}$ & This is a free-form text description of a goal identified from $P_i$'s intentions based on the actions in the actual story $S^a$.  We obtain $J=3$ goals from 3 annotators. & to play with their mom. \\
\hline
Goal Success  &  A label assignment for goal achievement in $S^a$ using the following choices: 1-Fully Successful, 2-Moderately Successful, 3-Success Unsure 4-Less Successful and 5-Unsuccessful.    &1-Fully Successful \\  
\hline 
\multicolumn{2}{c}{\textbf{Prospective Knowledge Annotation in Actual stories with HIT 1}} \\
\hline 
Next Action  &  This is a free-form text description of a likely next action involving the identified participant $P_i$ after the end of the story $S^a$. & They play with their mom.\\ \hline
Explanation  &  This is a free-form text description justifying the reason why the above next action is likely.  & They want to get their mom's attention and play with her.\\ \hline
Goal Direction  & A label assignment for whether the next action helps achieve the goal identified in $S^a$ using the following choices: 1-Fully Helps, 2-Somewhat Helps, 3-Unsure how it affects the goal 4-Does Not Help and 5-Contradicts with the goal.  &1-Fully Helps    \\ \hline
Goal Satisfaction  & A label assignment for whether the participant is likely to be happy with goal achievement after the events in $S^a$ and the next action  using the following choices: 1-Very Satisfied, 2-Moderately Satisfied, 3-Unsure 4-Less Satisfied and 5-Unsatisfied.  &4-Less Satisfied   \\  \hline
Goal Revision  & A label assignment for whether revising the participant's plan will help with goal achievement based on the outcome of events in $S^a$ and the next action, using the following choices: 1-Very Likely, 2-Somewhat Likely, 3-Unsure 4-Less Likely and 5-Unlikely.   & 1-Very Likely\\ \hline
Future Plan & This is a free-form text description of either an original plan or a revised plan to achieve the goal $G_{ij}$. & Marie should schedule `me time' when the kids are in bed for the night.\\
\hline 
\multicolumn{2}{c}{\textbf{Goal Inference Annotation in Alternative Stories  with HIT 2 }}\\
\hline 
Story Coherence   &  A label assignment indicating if the story makes sense using one of  ‘Does Not Make Sense’, ‘Somewhat Makes Sense’ and ‘Fully Makes Sense.’ & Fully Makes Sense\\ \hline
Incoherent Sentences & A list of sentences that make the story incoherent.  A story has only 5 sentences. & None\\  \hline
Goal inference & A label assignment for whether $P_i$ intends to achieve $G_{ij}$ in  $S^{c_k}$ 1-Inferable from Story, 0-Not Inferable from Story & 1- Inferable from Story.\\ \hline
Goal Success  & A label assignment for goal achievement in $S^{c_k}$ using the following choices: 1-Fully Successful, 2-Moderately Successful, 3-Success Unsure 4-Less Successful and 5-Unsuccessful.  &5-Unsuccessful  \\  
\hline 
\multicolumn{2}{c}{\textbf{Prospective Knowledge Annotation in Alternative Stories with HIT 3 }}\\
\hline 
Next Action Update   &  A binary decision of whether the next action needs to be updated for $S^{c_k}$. 1-Yes, 0-No &1-Yes\\ \hline
Explanation Update  &  A binary decision of whether the justification for the next action needs to be updated for $S^{c_k}$. 1-Yes, 0-No &1-Yes \\ \hline
Updated Next Action   &  This is a free-form text description of a likely next action involving the identified participant $P_i$ after the end of the story $S^{c_k}$.   & The kids fall a sleep. \\ \hline
Updated Explanation  &  This is a free-form text description justifying the reason why the  next action above for $S^{c_k}$ is most likely. &They are bored.\\ \hline
Goal Direction  & A label assignment for whether the next action annotated for  $S^{c_k}$ helps achieve the goal using the following choices: 1-Fully Helps, 2-Somewhat Helps, 3-Unsure how it affects the goal 4-Does Not Help and 5-Contradicts with the goal.      &5-Contradicts\\ \hline
Goal Satisfaction  & A label assignment for whether the participant is likely to be happy with goal achievement after the events in in $S^{c_k}$ and the next action  using the following choices: 1-Very Satisfied, 2-Moderately Satisfied, 3-Unsure 4-Less Satisfied and 5-Unsatisfied.    & 5-Unsatisfied\\  \hline
Goal Revision  & A label assignment for whether revising the participant's plan will help with goal achievement based on the outcome of events in $S^{c_k}$ and the next action, using the following choices: 1-Very Likely, 2-Somewhat Likely, 3-Unsure 4-Less Likely and 5-Unlikely.    &2-Somewhat Likely\\ \hline
Future Plan & This is a free-form text description of either an original plan or a revised plan to achieve the goal $G_{ij}$.  &The kids will cut the television cord when their mom is away at work.\\ \hline
\end{tabular}
}
\caption{A detailed description of Annotations collected with the various annotation HITs. 
}
\label{tab:HITS-annotation}
\vspace{-4mm}
\end{table*}

\subsection{HIT Designs}
\label{app:HIT_design}
This section provides supplementary details for the HIT design described in \cref{sec:HIT_design}. %
We developed 3 different HITs for collecting annotations for the actual story and the corresponding alternate stories.  As a reminder, each story can have up to 3 alternate stories.   In all three HITs, crowd workers are provided general annotation instructions along with a consent notice allowing them to leave the HIT if they choose not to annotate.  If they choose to annotate they are instructed to read a story displayed in the left column in a graphical format and follow instructions in the right column to produce a goal annotation set.  We also provide several annotated examples, highlighting both good and bad annotations for easy reference.

The actual story annotation set collected with the first HIT captures three types of inferred knowledge: 1) the selected participant's
overarching goal information as inferred from the story events,  2) anticipatory information of what happens next involving the participant after the story and whether it helps the participant with their goal achievement and 3) the participant's satisfaction level with their goal achievement and their subsequent plan, possibly involving a revision, to achieve their goal. The revision and plan are annotated only for goals that have not already been achieved in the story or later with the next action.  This includes goals that are not fully successful in the story and where the next action does not fully help achieve the goal. The revision annotation decides if the collected plan is a revised plan or an original plan.  If the next action does not help or contradicts with the goal achievement and a revision is not possible, the goal may not be achievable and we do not require a plan annotation (we do not force an annotator to provide a plan when one is not possible for the goal and story context).  Additionally, to understand annotator preference for the goal, we ask annotators to relate each story sentence to the goal using one of 6 \textbf{story-goal relations}: enabling the goal justifying the goal, blocking the goal, being the effect of an event in another sentence, being related to another event in another sentence but not related to the goal, being unrelated to the goal.  %
Relating this annotation to the example in \cref{fig:fig2}, the participant Marie's goal ``to enjoy some me time.'' is enabled by sentences 2 \& 3 and prevented by 4. We use these relation annotations during evaluation. 
\begin{figure*}[t!]
    \centering
\includegraphics[trim=0in 0.5in 0 .5in, scale=.5] 
    {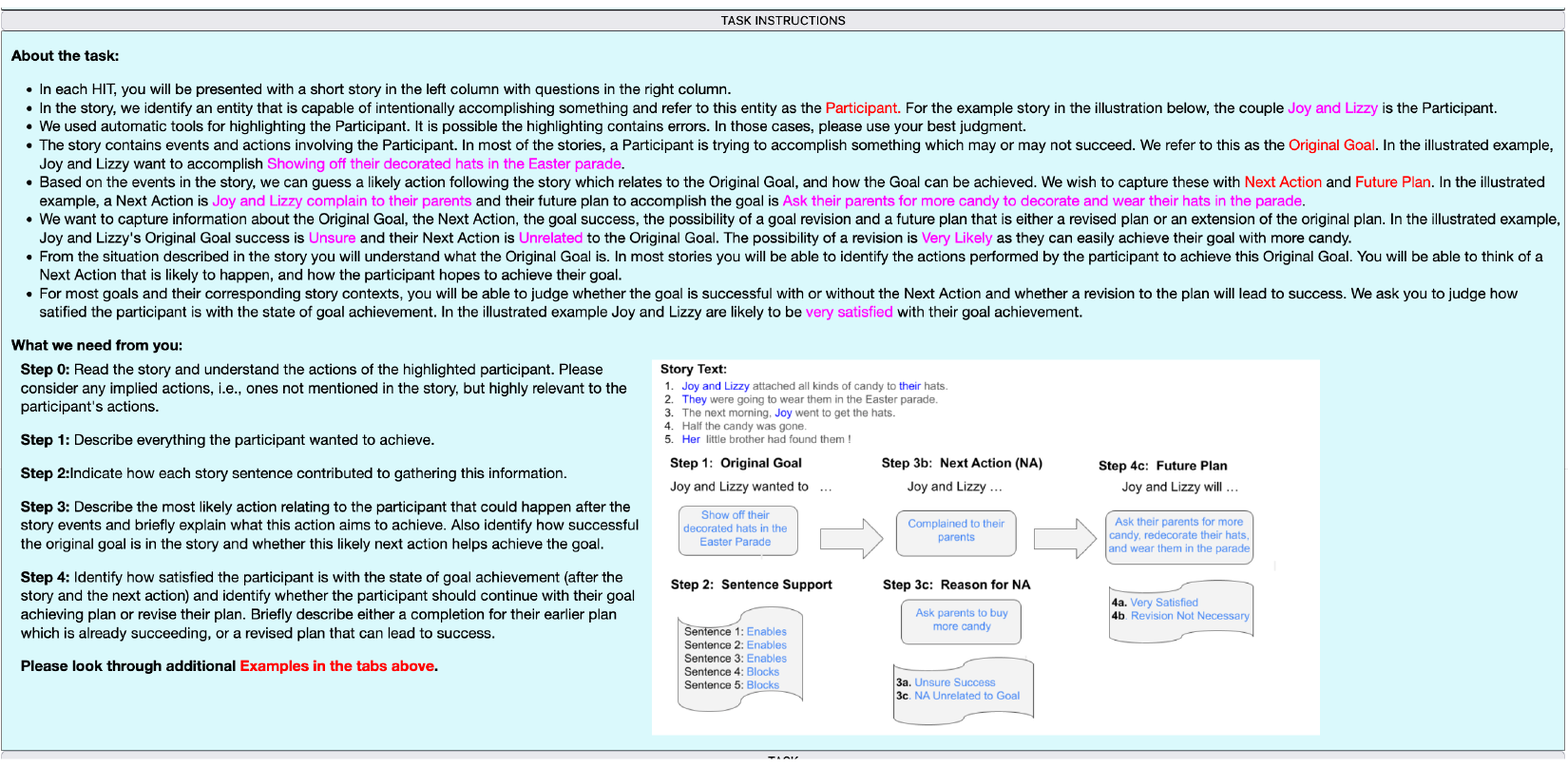}
	\caption{HIT General instructions for goal annotation in actual stories. %
 }
	\label{fig:screen_shot2}
\end{figure*}

\begin{figure*}[t!]
    \centering
\includegraphics[scale=.75, trim=0in 3in 0in 0in]{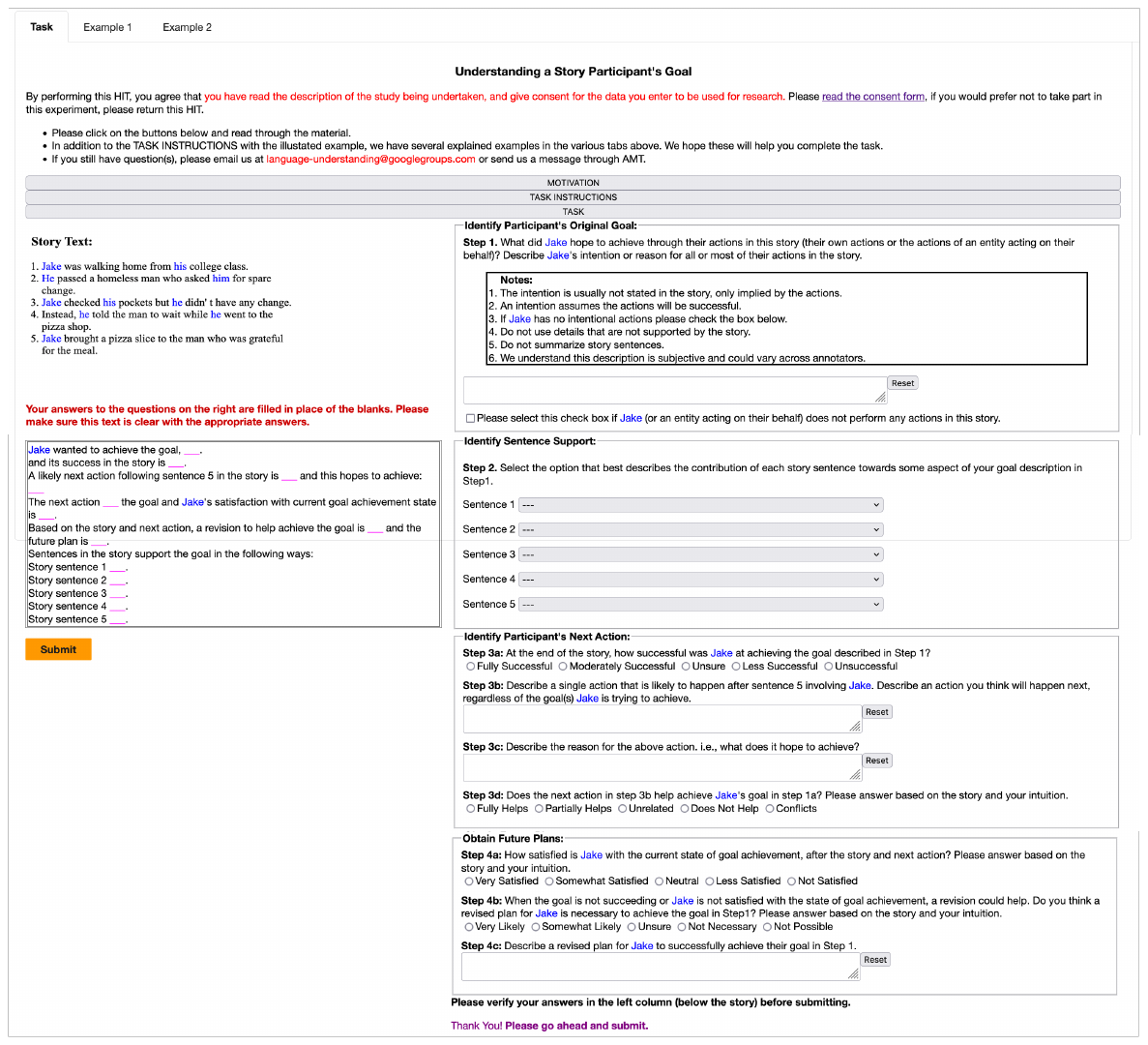}
	\caption{Screenshot of  HIT used in the goal annotation in actual stories.%
 }
	\label{fig:screen_shot1}
 \end{figure*}

The second HIT verifies two distinct aspects of an alternative story, $S^{c_k}$: whether $S^{c_k}$ makes sense and whether a selected participant still aims to achieve the goal (annotated for them in the $S^a$) in $S^{c_k}$.   We estimate that 10-15\% of the alternative stories in PASTA are under-specified and do not make sense from a goal annotation perspective (due to the constraints placed on implied states and story rewriting in PASTA). In this HIT, we present an alternative story and ask 3 crowd workers to identify if the story makes sense (selecting from 3 possible options: `Does Not Make Sense', `Somewhat Makes Sense' and `Fully Makes Sense') and select the story sentences that lead to the incoherence in the story when it does not `Fully Make Sense'. We keep track of this annotation to assess if story incoherence affects model performance in alternative situations (performance decrease is $<$.5\%).  

The selected participant's actions in the alternative story may no longer aim to achieve the goal (which they aimed to achieve in $S^a$).  To confirm if the goals annotated for $S^a$ are still valid in $S^{c_k}$, in this second HIT, we present the 3 goals collected for $S^a$ and ask which goals can be inferred from the participant's actions in $S^{c_k}$ and whether the inferred goals are achieved in the story. If all 3 annotators find that a goal cannot be inferred from $S^{c_k}$, we mark the story as not goal transferable and do not collect any further annotations.  When any of the workers annotates a goal as inferrable, we use another HIT to follow the achievement arc for the goal and obtain  further annotations.  The decision to gather goal annotations  even if a single worker identifies the goal as inferrable was based on the high IAA (96\% weighted fleiss-kappa) for the 3 workers on whether a goal can be inferred from the participant's actions.  However, we use the majority agreement for the gold label of goal transferability.

We use a third HIT when one or more annotators identify that a participant's goal is inferrable from $S^{c_k}$.  With this HIT we obtain a new set of goal annotations reusing and modifying the free-form text annotations from the actual story to obtain annotations that are also minimally updated reflecting the process used for obtaining the counterfactual in ~PASTA~\cite{ghosh2023pasta}. See \cref{tab:HITS-annotation} for the  annotations and the HITs used for obtaining them.

\subsection{Screenshots of Annotation HITS}
 
We present a few screen shots of the  
annotation HITs to show our general design. \Cref{fig:screen_shot2} shows the annotation instructions for this HIT, while \cref{fig:screen_shot1} shows the main goal annotation HIT. %
Our additional annotation and evaluation HITS are similar; in the interest of space utilization we are not including them here, though the templates are available with our released code.

\subsection{Crowd Sourcing Setup}
  
\subsubsection{HIT Streamlining}
\label{app:HIT_information:streamlining}
We ran several alpha runs of HITS (both for annotations and evaluations) and streamlined our instructions and examples 
until we were able get focused responses even with the subjectivity elicited by some of the annotations.  For example, what happens next after the story or a future plan to achieve the goal can lead to widely varying annotations.  This was observed in the evaluation IAA for both annotations and model generations. 

\subsubsection{Worker Selection and Qualifications}
\label{app:HIT_information:qualifications}
For our initial alpha runs of the HITs we used all workers who meet our community standard quality criteria, such as requiring a 98\% or greater HIT acceptance rate and the completion of 1000 approved HITs. In addition, we required the worker's stated location to be in the USA, UK, Canada, Australia, or New Zealand. We used the location requirement to avoid language-based artifacts given the language-dependent semantic phenomena and the high subjectivity our work can elicit.    
We did not use requester-generated qualification tests, though in early iterations we found that annotators who had completed at least 50 HITs in our prior work~\cite{poque} provided the most reliable annotations; the vast majority of our responses are from this group. %

 \subsubsection{Annotations and Pricing}
\label{app:compensation}
Our dataset contains 1024 participant-specific stories, each annotated by 3 workers. Of these 29 stories were discarded because one or more annotators identified the selected participants to be non-volitional.  For the remaining 995 stories and their corresponding 1060 counterfactual stories from PASTA, we obtained 6225 goal annotation sets.  Workers were paid an average of \$0.60 for each set of goal annotations for the actual story and an average of \$0.45 for each set of alternative annotations.  We targeted a pay of \$12-\$15 per hour and calculated HIT prices based on the average time spent by annotators on several test batches (we later verified the average time with actual annotation batches).  We believe the nature of this work 
allows us to obtain informative and generalizable knowledge in our subject area circumventing the typical positivity bias seen in AMT work~\cite{mturk_article}. 

\subsection{Evaluation and Quality Analysis} 
\subsubsection{Annotation Evaluation} 
\label{app:goal_eval}
The type of annotations and the nature of the stories in our dataset elicit a variation in the text style and format even from the same worker. Our experiments (\cref{sec:models}) did not uncover any easy biases attributable to the small number of workers selected for annotation. Since the second annotation HIT employed 3 crowd workers for the goal inference annotations, we report the IAA for the workers for these annotations in \cref{tab:evaluation-IAA}. 

\begin{table}[t]
\centering
\resizebox{.98\columnwidth}{!}{
\small
\begin{tabular}{|p{0.7\linewidth} | p{0.15\linewidth} | p{0.15\linewidth} |} %
\hline 
Evaluated Feature & IAA  & C+E \\
\hline\hline
\multicolumn{3}{|l|}{{\bf Goal Knowledge Evaluation}} 
\\ \hline 
Goal Coherence  &  81 \% & 86 \%  \\ \hline
Goal Explainability& 80 \% & 76 \%    \\ \hline
Goal Truthfulness   & 78 \% & 77 \% \\ \hline
Goal Faithfulness    & 74 \% & 65\%  \\ \hline
Goal Intentionality  & 74 \% & 77\%  \\ \hline
Goal Success in story & 84 \% & 78\%  \\  \hline
\hline
\multicolumn{3}{|l|}{{\bf Prospective Knowledge Evaluation in Original Stories}} 
\\ \hline
Next Action Cohesion  & 87\% & 80\%  \\ \hline
Next Action Coherence & 83\% & 85\%  \\ \hline
Next Action Explanation  & 85\% & 82\%  \\ \hline
Goal Direction  & 75\% & 72\%  \\ \hline
Goal Satisfaction & 81\% & 77\%  \\ \hline
Goal Revision  &  83\% & 74\% \\ \hline
Plan Correctness  & 83 \% & 74\%  \\ \hline
\hline
\multicolumn{3}{|l|}{\bf Goal Inference in Alternative Stories}
\\ \hline
Story Valid  & 78\% &  75\% \\ \hline
Goal Inference & 96\% &  95\% \\ \hline
Goal success in story  &  89\% & 85\%\\ \hline
\hline
\multicolumn{3}{|l|}{\bf Prospective Knowledge Eval. in Alternative Stories} 
\\ \hline
Next Action Cohesion  & 79\% & 77\%\\ \hline
Next Action Coherence & 76\% & 75\%\\ \hline
Next Action Explanation  & 79\% & 74\%\\ \hline
Goal Direction  & 69\%  & 62\%\\ \hline
Goal Satisfaction & 75\%  & 66\%\\ \hline
Goal Revision  & 76\%   & 75\%\\ \hline
Plan Correctness  & 77\%  & 74\%\\ \hline 
\end{tabular}
}
\caption{Evaluation of annotations from 100 actual stories and 209 alternative stories. Inter-rater Agreement scores using weighted Fleiss's Kappa~\cite{Marasini2016AssessingTI}. Average agreement is 80\%, which is quite high. %
}
\label{tab:evaluation-IAA}
\vspace{-5mm}
\end{table}

We evaluate the remaining annotations using 4 different evaluation HITs. 
The first evaluation HIT verifies the \textit{correctness} of the assigned relationship between each story sentence and the 3 goals obtained for each story.  With this HIT we evaluated 75 actual stories and the 225 annotated goals for these stories and verified that the described goals were overarching goals based on as many of the story sentences as possible.  The other 3 evaluation HITs evaluate the goal and prospective knowledge annotations in actual and alternative stories using criteria described in \cref{sec:dataeval}.  We also have an expert (the first author) evaluate 20\% of the evaluations and obtained a combined crowd and expert IAA.  
The quality of our goal annotations in actual stories is quite high with a crowd IAA above 80\% (except for the more subjective aspects of faithfulness to the story and participant intentionality).  We believe the difference between crowd and combined (crowd and expert) IAA is between 1-9\% shows that our crowd evaluations are of the expected quality.  Goal transferability to the alternative story with our crowd IAA of 96\% shows that our goal annotations for the actual story are of high quality and their applicability to the alternative story is well justified.  
The quality of our prospective knowledge annotations is quite high with an IAA above 80\% in actual stories and above 70\% in alternative stories (for all features except for goal direction because when a goal succeeded in the story the next actions and plan are subjective).

\subsubsection{Quality Analysis and Data Splits}
\label{app:quality-analysis}
For each annotation set, our evaluation using the above criteria resulted in scores for 13 features (not counting the evaluation of sentence features).  While the IAAs for all the evaluated features are good, we need a quality rating for an entire annotation set.  For this, we look to reasons for disagreement between the workers.  Since we evaluate 3 goals for each participant in a story, 12 workers comprehend each participant specific story either to annotate or to evaluate (3 annotators and 9 evaluators). We find that both nuances in semantic understanding and incorrect annotations lead to disagreements between the 12 workers.  So, we consider worker agreement in deciding the quality of an annotation set.  Annotation sets where each of the 13 features score $\geq$ 3.0 from a majority of annotators are considered to be of high quality.

We keep the stories in the train, dev and test splits mutually exclusive and use the actual story annotation set quality to decide the splits.  This is because goal descriptions were obtained from the actual stories and may not necessarily be a good fit for an alternative story. %
Not have a single high quality annotation set out of 12 possible ones (3 sets per single-participant story from 4 actual and alternate stories)  implies that the actual story is either complex or under-specified, leading to a difference in semantic understanding among the workers and low agreement on 1 or more features. %
Multiple participants in a story could also lead to low agreement, however, this was not the case; of the 14 stories and the 42 goal annotation sets that belong in this group all of them were either under-specified or had nuanced semantic meaning.  The  stories with at least one high quality story were randomly split in a 2:1 ration to create the test and validation splits respectively with all other annotated stories (including the ones identified as lower quality based on the agreement) making up the training split.  

Our quality assessment for assigning stories to data splits is very stringent requiring all 13 features to have a high agreement from a majority of evaluators.  However, we relax this requirement and heuristically apply one more filter to collect test and dev sets that are of good quality for both actual and alternative stories. We use features from story types for this process;  since we do not collect goal descriptions for an alternative story, we use the goal features from the actual story.  We allow 1 or 2 features out of the 13 to have a lower agreement when the overall average Likert score of all 13 features is $\geq$ 3.5. With this we discard any substandard evaluations that assign a Likert score of 3 for all annotations (approx. 7\%) but allow for some disagreement from nuanced stories (approx. 8\%).  We examined the nuanced stories and found that they belong to one of two types making it difficult to identify goal annotations: 1) the story consists of several participants where the selected participant has minimal actions.
2) the story consists of a single participant but latter story events substantially distract the participant from their original intended goal.
Overall, despite allowing some lower quality features, we employed a very strict requirement to ensure a high quality evaluation set. %
\begin{table}[t]
\centering
\resizebox{.98\columnwidth}{!}{
\small
\begin{tabular}{l||c|c||c|c|}
\hline 
& \multicolumn{2}{|c||}{Test Split} & \multicolumn{2}{|c|}{Dev Split}   \\
Evaluated Feature  & Actual & Alternate  & Actual & Alternate   \\
 \hline 
Goal Coherence  &  4.63 (98.6)&-&4.69 (100)&-\\ \hline
Goal Explainability& 4.61 (98.2)&-&4.67 (100)&-    \\ \hline
Goal Truthfulness   &4.72 (99.5)&-&4.74 (99.1)&-\\ \hline
Goal Faithfulness    &4.74 (99.1)&-&4.78 (100)&- \\ \hline
Goal Intentionality  & 4.42 (99.4)&-&4.61 (99.1)&-  \\ \hline
Goal Achievement in Story & 3.92 (81.3)&4.05 (84.6)&3.75 (75.5)&4.16 (87.9) \\  \hline
Next Action Cohesion  & 4.53 (97.7)&4.46 (96.8)&4.51 (100)&4.47 (98.1)\\ \hline
Next Action Coherence & 4.64 (99.1)&4.48 (96.6)&4.61 (100)&4.54 (98.1)    \\ \hline
Next Action Explanation  & 4.72 (100)&4.59 (97.5)&4.70 (99.1)&4.66 (99.5)  \\ \hline 
Goal Achievement with NA & 4.11 (84.9)&3.63 (70.5)&4.26 (88.7)&3.75 (74.3)  \\ \hline
Participant Satisfaction & 4.65 (99.5)&4.14 (89.5)&4.65 (97.2)&4.20 (90.3)  \\ \hline
Goal Revision  & 4.13 (89.1)&3.98 (89.1)&3.98 (91.9)&4.18 (93.8)   \\ \hline
Plan Correctness  & 4.13 (89.1)&4.01 (88.5) &3.98 (91.9)&4.15 (92.2)  \\ \hline
\end{tabular}
}
\caption{Quality of evaluation  splits using scores from 13 features. Average Likert scores of 3 crowd workers for all annotations in the data split are listed with the percentage of quality annotations (the average Likert score of the 3 workers for each annotation is $\geq 3.0$). The first 5 goal features are used for scoring both actual and alternate story annotations although they are obtained only for actual story goal annotations.}
\label{tab:splits-IAA}
\vspace{-5mm}
\end{table}

\subsection{Task Setup}
\label{app:task-setup}

\paragraph{Zero-shot vs. Few-shot} 
We use prompts similar to the RTE and WSC templates from the Flan-T5 templates collection~\cite{wei2022finetuned} which we list in \cref{tab:task-prompts} with examples.  We prompted models in a zero-shot and a few-shot setting with varying number of examples. 
While Flan-T5 model generations do not change between the two settings, the GPT-3.5 Turbo and GPT-4 models generate multi-sentence goals and next actions repeating story information along with an explanation in the zero-shot setting. We were unsuccessful at shortening the generations from these GPT models in  zero-shot setting even after trying a number of variations to the prompt including asking for a `concise', `short' or `brief' generation. In the few-shot setting, GPT models performance improves slightly with more number of examples however the input token limit of 1K tokens for Flan-T5  and 512 tokens T5 performance places a hard limit on the number of examples.  For an even comparison across all models we use a few-shot prompt setting with 3 examples for generative tasks and compared both settings for the NLI-type inferences.
We used 3 different random seeds and averaged the results (these differ by $<.01\%$). 
\paragraph{Fine-tuning} We found that fine-tuning on both actual and alternative annotations leads to better performance than fine-tuning on just the actual stories.  Additionally, 3-shot prompting a fine-tuned model improves performance significantly for some tasks (see task1 in \cref{sec:Tasks}).   
In our early development experiments, we used 5-fold cross validation training for 5 to 10 epochs.   Noticing that evaluation loss plateaus in 3 to 5 epochs depending on the model type, we stop training after 5 epochs, saving a checkpoint at each epoch.  We train all the training data and evaluate with both the validation and test splits using the checkpoint at the 3rd epoch for generative tasks and goal applicability and epoch 4 for the other NLI type tasks.  Results reported are an average of 3 model runs with different initial random seeds of 4, 7 and 11 (the variance in results across the various scores is $<.01\%$).   
 
\paragraph{Infrastructure}
We trained our models we used both RTX 8000 with 48GB of GPU memory and Nvidia A100 with 80GB GPU memory. Approximate run time for a model is a less than 30 minutes.

\paragraph{Hyperparameters}
For all experiments we used AdamW~\cite{DBLP:journals/corr/abs-1711-05101} optimizer, a learning rate of $10-4$, a weight decay of $10^-4$ and 3 different random seeds of 4, 7 and 11.   
We applied manual tuning and tried various learning rates from .001 to .00001 as suggested for T5 models.  For the generation we used  Top-K sampling with a beam size of 2 or a 3-shot prompt setting.  These parameters worked well for all the models and were selected based on the initial tests across all tasks.

\subsubsection{Additional Metrics and Models }
\label{app:additional-metrics-models}

We present additional automated metrics (Rouge1, Rouge2, RougeL, BERTScore, Corpus and Google's version of Sentence BLEU in this section with additional models). We note that given the wide variety of possible wordings, BLEU is not well suited for this type of generation. While this is a known issue with generation involving deeper natural language understanding, these results highlight the shortcomings, and provide strong evidence for future work to continue examining how to effectively automatically evaluate generated natural language.

\begin{table}[t]
 \centering
\resizebox{.98\columnwidth}{!}{
 \begin{tabular}{|l|c|c|c|c|c|}
 \hline
 &\multicolumn{5}{|c|}{\textbf{Average Likert Scores (\# evaluations with score $\geq$ 3.0)}}\\
Model Type  &Coherence&Explainable&Faithful&Truthful&Intentional\\ 
\hline

Ref Avg ($\sigma^2 <.1$)  &4.68 (25) &4.67 (25) &4.73 (25) &4.80 (25) &4.70 (25) \\ 
 fT5b   (3-shot) &\underline{3.79} (42)&\underline{3.84} (20)&\underline{4.08} (21) &\underline{3.69} (22)&\underline{3.64} (18) \\
 fT5xxl (3-shot)&4.47 (24)&4.40 (23)&4.76 (25)&4.71 (25)&\underline{4.28} (22) \\
 gpt3.5t (3-shot)&4.53 (24)& 4.83 (25)&4.85 (24)&4.83 (24)&4.32 (23) \\
 gpt4  (3-shot) &4.77 (25)&4.77 (25)&4.85 (25)&4.76 (25)&4.68 (24)\\
\hline
\end{tabular}
}
 
\caption{\textbf{Task 1} dev results: Human Evaluation of model generated goals (for volitional participants in \textbf{actual stories from the dev split}). See \cref{sec:model_eval_metrics} \&  \cref{sec:goal_eval}  for evaluation details;  \cref{tab:goal-automated-metrics} for additional models and automated metrics.}
\label{tab:goal-manual-metrics-dev}

\end{table}

\begin{table}[t]
 \centering
\resizebox{.98\columnwidth}{!}{
 \begin{tabular}{|l|c|c|c|c|}
\hline
Model& Overall & Full Agreement& Partial Agree. \\
\hline
Majority (0/1 labels)   &.74 (.31/.74)& .86 (.17/.86) &.00 (.17/.00) \\
T511b (3-shot) &.48 (.29/.53) &.51 (.17/.57) &.50 (.54/.43) \\
fT5b (0- \& 3-shot) &.74 (.29/.84) &.83 (.28/.89) &.44 (.30/.59)\\
ft5l (0- \& 3-shot)  &.77 (.27/.88) &.88 (.30/.94) &.43 (.25/.62)\\
ft5xl (0- \& 3-shot)  &.61 (.43/.65) &.65 (.29/.69) &.54 (.69/.38)\\
fT5xxl (0- \& 3-shot)&.81 (.60/.86) & .86 (.50/.89) &.65 (.71/.58)\\
gpt3.5t (0-shot)&.73 (.49/.78) &.79 (.39/.84) &.50 (.72/.61)\\
gpt3.5t (3-shot)&.76 (.53/.82) &.81 (.41/.86) &.60 (.67/.52)\\
gpt4  (0-shot)&.80 (.59/.85) &.84 (.48/.88) &.67 (.72/.61)\\
gpt4  (3-shot)&.80 (.59/.85) & .85 (.48/.88) &.65 (.72/.58)\\
\hline
T5b-ft (0- \& 3-shot)&.80 (.39/{.89}) & {.89} (.39/{.94}) &.53 (.29/{.67})\\
fT5b-ft (0- \& 3-shot)&{.81} (.44/{.89}) &{.89} (.46/{.94}) &.50 (.41/{.67})\\
fT5l-ft (0- \& 3-shot)&{.89 (.65/.94)}&{.95 (.71/.97) }&{.68 (.61/.65)}\\
fT5xl-ft (0- \& 3-shot)&{.92 (.76/.96)}&{.97 (.84/.99)}&{.75 (.70/.80)}\\
\hline
\end{tabular}}
 \caption{\textbf{Task 2} results: Goal transferability comparing performance for full and partial agreement in \textbf{alternative stories} (test split) using \textbf{weighted F1} and (F1 for not/yes transferable).  
}
\label{tab:goal-applicability-metrics-more}
\vspace{-5mm}
\end{table}

\begin{table}[t]
 \centering
\resizebox{.98\columnwidth}{!}{
 \begin{tabular}{|l|c|c|c|c|c|c|}
 \hline
&\multicolumn{6}{|c|}{\textbf{Average Likert Scores (\# evaluations with score $\geq$ 3.0)}}\\
& \multicolumn{3}{|c|}{\textbf{Actual Stories}} & \multicolumn{3}{|c|}{\textbf{Alternative Stories}}\\ 
 Model Type & Coherence &Cohesion & Explain. & Coherence &Cohesion & Explain.\\
\hline
 Ref     &4.55 (25)&4.38 (25)&4.69 (25)&4.51 (24) &4.48 (24)&4.34 (25)\\
 \hline
 fT5b (3-shot) &\underline{2.11} (2)& \underline{1.86} (1) & \underline{1.39} (0) &\underline{2.14} (2)&\underline{2.08} (3)&\underline{1.86} (1)\\
 fT5xxl (3-shot)&4.63 (25)&4.44 (23)&\underline{4.13} (21)&3.82 (25)&\underline{3.71} (25)&\underline{3.58} (19)\\
 gpt3.5t (3-shot)&4.67 (25)&4.63 (25) &4.83 (25)&4.65 (25)&4.60 (25)&4.79 (25)\\
 gpt4 (3-shot) &4.76 (25)&4.72 (25)&4.81 (25)&4.75 (25)&4.73 (25)&4.80 (25)\\
 \hline
\end{tabular}
}
 \caption{\textbf{Task 3a} dev results: Evaluation of Next Actions with Explanations (evaluated generations contained both) \textbf{for stories from the dev split}. 
 Scores underlined  are significantly lower than reference.  See \cref{sec:model_eval_metrics} \&  \cref{sec:goal_eval}  for evaluation details;  See  \cref{tab:next-action-automated-metrics}  for additional models and automated evaluation metrics.%
 }
\label{tab:next-actions-manual-metrics-dev}
\end{table}

\begin{table}[t]
 \centering
\resizebox{.98\columnwidth}{!}{
 \begin{tabular}{|l|c|c||c|}
\hline
&\multicolumn{2}{|c||}{\textbf{Task 3b:  Next Action}} &\multicolumn{1}{|c|}{\textbf{Task 3c:  Explanation}} \\
Model Type & macroF1 & wtF1 (Un-/Most/Uns F1)  & wtF1  (0/1 F1)\\ 
\hline
Maj.  &.48&.91 (.28/.12/.94)&.57 (.30/.82)\\
\hline
T511b (3-shot) &.31&.53 (.25/.00/.70) &.52 (.33/.60) \\
fT5b (0- \& 3-shot) &.30&.55 (.12/.00/.78) &.70 (.47/.79)\\
fT5l (0- \& 3-shot)  &.37&.60 (.37/.00/.74)&.78 (.64/.84)\\
fT5xl (0- \& 3-shot)  &.43&.66 (.52/.00/.77)&.73 (.65/.76)\\
fT5xxl (0- \& 3-shot)&.45&.67 (.59/.00/.76)&.79 (.69/.83)\\
gpt3.5t (0-shot)&.33&.47 (.44/.04/.51)&.62 (.45/.70)\\
gpt3.5t (3-shot)&.37&.48 (.44/.13/.52)&.59 (.40/.67)\\
gpt4  (0- \& 3-shot)&.47&.72 (.59/.00/.83)&.84 (.71/.90)\\
\hline
T5b-ft (0- \& 3-shot)&.42&.65 (.47/.00/.77)&.68 (.42/.79) \\
fT5b-ft (0- \& 3-shot)&.43&.67 (.51/.00/.78)&.70 (.49/.79)\\
fT5l-ft (0- \& 3-shot)&{.49}&{74} ({.63}/.00/.83)&.75 (.57/.83)\\
fT5xl-ft (0- \& 3-shot)&{.53}&{.75} ({.65}/.11/{.84})&.79 (.64/.85)\\
\hline
\end{tabular}
}
 \caption{\textbf{Task 3b, 3c} results: Next actions and Explanations transferability to alternative stories (\textbf{test split}).  
}
\label{tab:next-action-applicability-metrics-more}
\end{table}

 We present results for GPT-3.5 Turbo, GPT-4, T5-11b and various Flan-T5 model sizes (base, large, XL and XXL), using both 0-shot and 3-shot prompt settings. Additionally, we present results for T5-base, Flan-T5 base, large and XL models fine-tuned on the SAGA dataset.  These results are as follows:
 \begin{enumerate}
\item In \cref{tab:goal-manual-metrics-dev} we present additional results for task 1 (human evaluation of model generated goals) for the dev split. As noted in the main paper, larger models perform well on this data split.  See \cref{task1} in the main paper for a detailed discussion of models performance.  

\item In \cref{tab:goal-applicability-metrics-more} we list the macro F1 scores and F1 scores for both the positive and negative labels along with a weighted F1 for identifying goal applicability.  Most models do well on identifying positive labels in the full agreement setting, but struggle with the negative labels.  We discuss the F1 results for the negative labels in the main paper in \cref{task2}.  In the partial agreement setting, larger models perform similarly for both label settings.

\item In \cref{tab:next-actions-manual-metrics-dev} we present results from the human evaluation of model generated next actions for the dev split. As noted in the main paper, larger models perform well on this data split. Flan-T5-XXL generations contain similar issues as previously discussed with the test data.   See \cref{task3a} for a detailed discussion on these issues.  

\item In \cref{tab:next-action-applicability-metrics-more} we present macro F1 and the F1 scores for the individual labels along with the weighted F1 scores for the transferability of Next Actions and Explanations from the actual stories to alternative stories.  We note most models except GPT-3.5 Turbo and fine-tuned Flan-T5-XL are unable to identify the `Unsure' labels for the Next Action transferability inference.  All models are better at identifying the positive labels than negative labels for Explanation transferabilty inference.  See \cref{task3b} in the main paper for a detailed discussion.

\item In \cref{tab:in-story-metrics}, \ref{tab:after-story-metrics} and \ref{tab:after-story-metrics-more} we present macro F1 and the F1 scores for the individual labels along with the weighted F1 scores for identifying achievement in the story, achievement after the story with the next action and participant's satisfaction towards goal achievement.  We present results for both 0-shot and 3-shot prompting and show that it leads to improvement in most models. A few exceptions are where we see a decrease of 1-2\% performance with 3-shot prompting in the larger models for identifying achievement after the story and identifying participant's satisfaction.   See \cref{task5} in the main paper for a detailed discussion.

\item In \cref{tab:goal-automated-metrics}, \ref{tab:next-action-automated-metrics} and \ref{tab:future-plan-automated-metrics} we report the automated metrics for model generations of goal descriptions (Task 1), next actions (Task 3a) and future plans (Task 4).

\item In \cref{tab:model-goal-generations}, \ref{tab:model-action-generations} and \ref{tab:model-plan-generations} we list a few example generations and identify some of the issues.  T5-11b generations are only listed in \cref{tab:model-goal-generations} to show the additional generated text along with the goal description.  We do not list T5-11b generations for the next actions and future plans for better utilization of space.

\item In \cref{tab:task-prompts} we list the prompts used in our tasks.
 \end{enumerate}

\begin{table}[t]
 \centering
\resizebox{.98\columnwidth}{!}{
 \begin{tabular}{|l|c|c||c|c|}
 \hline
& \multicolumn{2}{|c||}{\textbf{Actual Stories}} &  \multicolumn{2}{|c|}{\textbf{Alternative Stories}}\\ 
 \cline{2-5}
& macroF1  & wt. F1 ( 0/1/Uns F1)   & macroF1  & wt. F1 ( 0/1/Uns-F1)   \\ 
\hline
 Maj.   &.38& (.25/.85/.20)& .22& (.61/.65/.15)\\ 
 T511b (0- \& 3-shot) &.38&.63 (.35/.78/.00)&.35&.49 (.46/.60/.00) \\
 fT5b (0- \& 3-shot)  &.28&.63 (.00/.85/.00)&.32&.45(.29/.67/.00)\\
 fT5l (0- \& 3-shot)  &.43&.67 (.48/.81/.00)&.49&.67(.72/.74/.00)\\
 fT5xl (0- \& 3-shot) &.46&.70(.54/.83/.00)&.55&.75(.82/.83/.00)\\
 fT5xxl (0- \& 3-shot)&.47&.71(.57/.85/.00)&.55&.76(.83/.83/.00)\\
 gpt3.5t(0-shot)      &.41&.65(.45/.79/.00)&.51&.70(.76/.78/.00)\\
 gpt3.5t(3-shot)      &.42&.66(.46/.79/.00)&.54&.71(.76/.77/.10)\\
 gpt4 (0-shot)        &.45&.64(.54/.75/.00)&.66&.80(.85/.83/.30)\\
 gpt4 (3-shot)        &.46&.66(.53/.77/.00)&.68&.85(.84/.35/.35)\\
 \hline
 T5b-ft (0- \& 3-shot)&.40&.67(.37/.83/.00)&.52&.69(.75/.76/.05)\\
fT5b-ft (0- \& 3-shot)&.44&.69(.46/.83/.00)&.50&.67(.72/.75/.00)\\
fT5l-ft (0- \& 3-shot)&.45&.70(.52/.84/.00)&.53&.71(.76/.77/.05)\\
fT5xl-ft (0- \& 3-shot)&.47&.71(.57/.86/.00)&.56&.71(.75/.78/.04)\\
 \hline 
\end{tabular}
}
 \caption{\textbf{Task 5a} results: Model identification of goal achievement within the story (in both story types).  
 }
\label{tab:in-story-metrics}
\end{table}

\begin{table}[t]
 \centering
\resizebox{.98\columnwidth}{!}{
 \begin{tabular}{|l|c|c||c|c|}
 \hline
& \multicolumn{2}{|c||}{\textbf{Actual Stories}} &  \multicolumn{2}{|c||}{\textbf{Alternative Stories}} \\
\cline{2-5}
Model Type& macroF1  & wt. F1 & macroF1  & wt. F1    \\ 
\hline
 Maj. (/0/1/Un)   &.28&(.47/.79/.09)& .27& (.48/.79/.06)\\ 
 T511b (3-shot)  &.33&.54(.47/.61/.00)&.32&.46(.40/.51/.00)\\
 fT5b (0-shot)   &.28&.52(.05/.78/.00)&.30&.54(.12/.77/.00)\\
 fT5b (3-shot)   &.41&.66(.47/.80/.00)&.33&.54(.27/.69/.00)\\
 fT5l (0-shot)   &.52&.77(.72/.85/.00)&.33&.47(.52/.47/.00)\\
 fT5l (3-shot)   &.53&.77(.73/.85/.00)&.34&.49(.53/.50/.00)\\
 fT5xl (0-shot)  &.56&.81(.80/.87/.00)&.27&.35(.51/.30/.00)\\
 fT5xl (3-shot)  &.56&.83(.81/.89/.00)&.28&.37(.49/.33/.00)\\
 fT5xxl (0-shot) &.58&.86(.86/.91/.00)&.29&.40(.87/.37/.00)\\
 fT5xxl (3-shot) &.59&.83(.83/.89/.00)&.32&.46(.50/.47/.00)\\
 gpt3.5t(0-shot) &.60&.79(.80/.82/.17)&.27&.35(.40/.33/.00)\\
 gpt3.5t(3-shot) &.58&.84(.83/.90/.00)&.44&.66(.49/.72/.00)\\
 gpt4 (0-shot)   &.43&.64(.54/.74/.00)&.49&.74(.67/.80/.00)\\
 gpt4 (3-shot)   &.41&.62(.52/.70/.00)&.56&.83(.78/.90/.00)\\
 \hline
 T5b-ft (0-shot) &.50&.76(.68/.85/.00)&.42&.64(.54/.72/.00)\\
 fT5b-ft (0-shot)&.55&.76(.67/.86/.00)&.44&.68(.55/.77/.00)\\
 fT5l-ft (0-shot)&.58&.86(.79/.92/.00)&.46&.70(.62/.77/.00)\\
 fT5xl-ft (0-shot)&.59&.85(.86/.91/.00)&.52&.71(.63/.78/.00)\\
 \hline
\end{tabular}
}
 \caption{\textbf{Task 5b} results: Models' identification of  goal achievement with next action in both story types (\textbf{test split}).  
 }
\label{tab:after-story-metrics}
\vspace{-5mm}
\end{table}

\begin{table}[t]
 \centering
\resizebox{.98\columnwidth}{!}{
 \begin{tabular}{|l|c|c||c|c|}
 \hline
& \multicolumn{2}{|c||}{\textbf{Actual Stories}} &  \multicolumn{2}{|c|}{\textbf{Alternative Stories}} \\
\cline{2-5} 
Model Type& macroF1  & wt. F1 & macroF1  & wt. F1 \\ 
\hline
Maj.   &.28& (.35/.85/.10) &.26& (.45/.66/.17)\\ 
 T511b (3-shot)  &.33&.61(.22/.76/.00)&.31&.46(.22/.64/.00)\\
 fT5b (0-shot)   &.30&.64(.04/.85/.00)&.35&.56(.28/.78/.00)\\
 fT5b (3-shot)   &.35&.66(.18/.85/.00)&.36&.56(.32/.76/.00)\\
 fT5l (0-shot)   &.50&.78(.60/.88/.00)&.44&.62(.57/.74/.00)\\
 fT5l (3-shot)   &.47&.74(.57/.85/.00)&.40&.56(.53/.65/.00)\\
 fT5xl (0-shot)  &.53&.82(.68/.92/.00)&.45&.64(.61/.75/.00)\\
 fT5xl (3-shot)  &.51&.79(.65/.88/.00)&.42&.59(.57/.68/.00)\\
 fT5xxl (0-shot) &.54&.83(.71/.92/.00)&.45&.65(.63/.76/.00)\\
 fT5xxl (3-shot) &.52&.79(.65/.89/.00)&.46&.63(.63/.73/.00)\\
 gpt3.5t(0-shot) &.37&.55(.44/.61/.07)&.33&.41(.51/.41/.06)\\
 gpt3.5t(3-shot) &.53&.81(.68/.91/.00)&.47&.61(.60/.69/.12)\\
 gpt4 (0-shot)   &.50&.75(.68/.82/.00)&.44&.58(.61/.64/.07)\\
 gpt4 (3-shot)   &.56&.85(.76/.93/.00)&.49&.66(.65/.77/.05)\\
 \hline
 T5b-ft (0-shot) &.63&.86(.74/.93/.22)&.52&.69(.62/.81/.11)\\
 fT5b-ft (0-shot)&.65&.85(.74/.92/.30)&.51&.66(.58/.76/.19)\\
 fT5l-ft (0-shot)&.60&.85(.70/.94/.17)&.52&.70(.62/.82/.13)\\
 fT5xl-ft (0-shot)&.65&.86(.73/.94/.13)&.75&.74(.71/.84/.13)\\
 \hline
\end{tabular}
}
 \caption{\textbf{Task 5c} results: Models' identification of participants' satisfaction towards goal achievement in both story types (\textbf{test split}).  
 }
\label{tab:after-story-metrics-more}
\vspace{-5mm}
\end{table}

\begin{table*}[t]
 \centering
\resizebox{.98\textwidth}{!}{
 \begin{tabular}{|l|c|c|c|c|c|c|c||c|c|c|c|c|c|c|}
 \hline
       & \multicolumn{7}{|c||}{Actual Stories} & \multicolumn{7}{|c|}{Alternative Stories}\\
 Model & \multicolumn{3}{|c|}{Rouge} & Met & \multicolumn{2}{|c|}{BLEU} & Bert & \multicolumn{3}{|c|}{Rouge} & Met & \multicolumn{2}{|c|}{BLEU}&Bert\\ 
  & R1 & R2 & RL & eor & Cor. & Sen.&Score&R1 & R2 & RL & eor & Cor& Sen.&Score\\ 
\hline
 \multicolumn{15}{|c|}{Test Data Split}\\
\hline
 T5-11b      & .16 & .07 & .14 & .28 & .02 & .03 & .86 & .16 & .06 & .14 & .26 & .02 & .03 &.86 \\
 flanT5b     & .43 & .21 & .42 & .40 & .12 & .16 & .86 & .43 & .19 & .41 & .39 & .10 & .15 &.86 \\
 flanT5l     & .45 & .24 & .45 & .33 & .18 & .15 & .86 & .44 & .20 & .43 & .31 & .16 & .14 &.86 \\
 flanT5xl    & .48 & .25 & .46 & .36 & .17 & .15 & .86 & .44 & .22 & .43 & .33 & .15 & .17 &.86 \\
 flanT5xxl   & .52 & .29 & .50 & .49 & .21 & .22 & .86 & .47 & .26 & .46 & .44 & .18 & .19 &.86 \\
 gpt3.5t     & .45 & .25 & .44 & .52 & .13 & .17 & .86 & .44 & .23 & .42 & .48 & .11 & .15 &.86 \\
 gpt4        & .55 & .33 & .54 & .55 & .23 & .26 & .86 & .49 & .27 & .48 & .49 & .19 & .22 &.86 \\
 T5b (ft)    & .34 & .15 & .33 & .26 & .07 & .17 & .91 & .32 & .14 & .31 & .25 & .06 & .14 &.91 \\
 flanT5b (ft)& .33 & .17 & .33 & .23 & .08 & .19 & .91 & .33 & .17 & .32 & .24 & .09 & .20 &.92 \\
 flanT5l (ft)& .37 & .18 & .36 & .27 & .10 & .25 & .92 & .37 & .18 & .36 & .27 & .10 & .22 & .92 \\
\hline
\multicolumn{15}{|c|}{Dev Data Split} \\
\hline
 T5-11b      & .15 & .07 & .14 & .26 & .03 & .03 & .86 & .15 & .06 & .13 & .26 & .02 & .03 & .86 \\
 flanT5b     & .39 & .16 & .38 & .36 & .09 & .14 & .86 & .35 & .11 & .34 & .31 & .07 & .12 & .86 \\
 flanT5l     & .43 & .24 & .42 & .31 & .18 & .15 & .86 & .36 & .16 & .35 & .26 & .11 & .12 & .86 \\
 flanT5xl    & .44 & .23 & .43 & .35 & .16 & .18 & .86 & .41 & .19 & .39 & .31 & .14 & .15 & .86 \\
 flanT5xxl   & .45 & .26 & .44 & .44 & .19 & .20 & .86 & .44 & .23 & .42 & .41 & .17 & .17 & .86 \\
 gpt3.5t     & .45 & .26 & .43 & .50 & .17 & .14 & .86 & .42 & .22 & .40 & .46 & .15 & .12 & .86 \\
 gpt4        & .51 & .30 & .48 & .53 & .23 & .20 & .86 & .47 & .28 & .46 & .49 & .23 & .21 & .86 \\
 T5b (ft)    & .32 & .14 & .31 & .25 & .16 & .07 & .91 & .32 & .12 & .30 & .23 & .13 & .05 & .91 \\
 flanT5b (ft)& .29 & .13 & .28 & .21 & .15 & .06 & .92 & .31 & .13 & .29 & .22 & .15 & .06 & .92 \\
 flanT5l (ft)& .34 & .17 & .34 & .25 & .19 & .08 & .92 & .34 & .17 & .34 & .26 & .20 & .08  &.92 \\
 \hline
 \end{tabular}
 }
 \caption{\textbf{Task 1} results: Automated Evaluation scores for model generated goals  for volitional participants in \textbf{actual and alternative stories} from the \textbf{test and dev splits}.  
 }
\label{tab:goal-automated-metrics}
\vspace{-7mm}
\end{table*}

\begin{table*}[t]
\centering
\resizebox{.98\textwidth}{!}{
\small
\begin{tabular}[trim=0 .2in 0 0,  width=.95\linewidth]{p{0.30\linewidth} |p{0.15\linewidth} |p{0.50\linewidth} }
\hline\hline
Story &  Model & Goal\\
\hline 
\multirow{7}{=}{\textbf{[Actual] Story:} Jared was tossing and turning all night. He kept hearing a buzz right next to his ear. Every time he tried to swat, the sound went away.  As he was drifting off, something landed right on top of his ear. He hit it and discovered that it was a now dead fly! [actual story]} 
 & Reference & To sleep peacefully\\
 &T511b (3-shot) & ''''''':iiiii'''':iiii:: Jared wanted to get rid of the buzzing in his ear. Story: Jared was trying to sleep. He kept hearing a buzzing sound. He kept swatting at it. It kept buzzing. He was getting frustrated.\\
&fT5b (3-shot)&  To get rid of the fly.\\
&fT5l (3-shot)& swat the fly\\
&fT5xl (3-shot)& Jared wanted to sleep.\\
&fT5xxl (3-shot)& To get rid of the buzzing sound. \\
&gpt3.5t (3-shot)& Get rid of the annoying fly and be able to sleep peacefully. \\
&gpt4 (3-shot)& ,Get rid of the disturbance (the fly) so he could sleep.\\
&T5b-ft (0-shot)& to catch a fly\\
&fT5b-ft (0-shot)&toss and turn all night.\\
&fT5b-ft (3-shot)&to get rid of the buzz \textbf{[3-shot prompting leads to a better goal description than the 0-shot prompting]}\\
 \hline
\multirow{7}{=}{\textbf{[Alternative] Story:} Martha has never liked wearing deodorant. However, her students notice that she has bad body odor. One day they tell her she needs to start wearing deodorant.  Martha tries deodorant and is surprised that it isn't that bad. Now she doesn't mind wearing deodorant at all. } 
& Reference & Smell less bad around her classmates. \\
&T511b (3-shot)& ::::::::::::::::::::: Put out the fire. Story: Paul jumped out of his car and locked the door. He jogged into the building and locked the door. He mentally prepared himself. He::odor:: Start wearing deodorant.\\
&fT5b (3-shot)& To be a good student\\
&fT5l (3-shot)& wear deodorant\\
&fT5xl (3-shot)& She wants to smell nice\\
&fT5bxxl (3-shot)& She wants to smell good\\
&gpt3.5t (3-shot)& to improve her personal hygiene and prevent body odor\\
&gpt4 (3-shot) & Eliminate her body odor\\
&T5b-ft & wear deodorant to get rid of her bad body odor \\ 
&fT5b-ft (0-shot)&Wear deodorant\\
&fT5b-ft (3-shot)&Change her body odor.
\textbf{[3-shot prompting leads to a better goal description than the 0-shot prompting]}\\
\hline 
\hline
\end{tabular}
}
\caption{\textbf{Task 1} examples:  Model generated Goals.  T5-11b model generates additional text either before or after the goal.  Issues with generations are described in bold in square brackets.
}
\label{tab:model-goal-generations}
\vspace{-4mm}
\end{table*}

\begin{table*}[t]
 \centering
\resizebox{.98\textwidth}{!}{
 \begin{tabular}{|l|c|c|c|c|c|c|c||c|c|c|c|c|c|c|}
 \hline
       & \multicolumn{7}{|c||}{Actual Stories} & \multicolumn{7}{|c|}{Alternative Stories}\\
 Model & \multicolumn{3}{|c|}{Rouge} & Met- & \multicolumn{2}{|c|}{BLEU}&Bert & \multicolumn{3}{|c|}{Rouge} & Met- & \multicolumn{2}{|c|}{BLEU}&Bert\\ 
  & R1 & R2 & RL & eor & Cor. & Sen.&Score&R1 & R2 & RL & eor & Cor. & Sen.&Score\\ 
\hline
\multicolumn{15}{|c|}{Test Data Split}\\
\hline
 T5-11b       & .34 & .22 & .30 & .42 & .13 & .14 & .86 & .35 & .22 & .30 & .41 & .14 & .15 & .86\\
 flanT5b      & .32 & .20 & .31 & .31 & .21 & .21 & .86 & .32 & .19 & .31 & .31 & .20 & .20 & .86\\
 flanT5l      & .15 & .04 & .13 & .10 & .02 & .05 & .86 & .15 & .04 & .14 & .10 & .02 & .06 & .86\\
 flanT5xl     & .26 & .16 & .25 & .23 & .13 & .16 & .86 & .26 & .16 & .24 & .23 & .12 & .15 & .86\\
 flanT5xxl    & .50 & .37 & .47 & .51 & .35 & .33 & .86 & .50 & .37 & .46 & .49 & .34 & .33 & .86\\
 gpt3.5t      & .46 & .32 & .43 & .48 & .25 & .26 & .86 & .45 & .30 & .41 & .47 & .24 & .25 & .86 \\
 gpt4         & .50 & .36 & .46 & .51 & .32 & .31 & .86 & .49 & .35 & .45 & .49 & .32 & .30 & .86 \\
 T5b (ft)     & .51 & .36 & .47 & .49 & .29 & .37 & .90 & .51 & .36 & .47 & .47 & .28 & .37 & .91\\
 flanT5b (ft) & .50 & .36 & .48 & .49 & .30 & .39 & .90 & .51 & .38 & .49 & .49 & .29 & .42 & .91\\
 flanT5l (ft) & .52 & .38 & .49 & .50 & .30 & .41 & .92 & .52 & .38 & .49 & .49 & .27 & .41 & .91\\
\hline
\multicolumn{15}{|c|}{Dev Data Split}\\
\hline
 T5-11b       & .33 & .22 & .29 & .40 & .13 & .12 & .86 & .33 & .21 & .29 & .40 & .14 & .13 & .86\\
 flanT5b      & .31 & .20 & .29 & .29 & .19 & .20 & .86 & .33 & .20 & .31 & .31 & .20 & .20 & .86\\
 flanT5l      & .16 & .04 & .15 & .10 & .05 & .02 & .86 & .17 & .05 & .15 & .12 & .07 & .03 & .86 \\
 flanT5xl     & .26 & .15 & .24 & .22 & .15 & .13 & .86 & .25 & .17 & .24 & .23 & .16 & .12 & .86\\
 flanT5xxl    & .47 & .35 & .44 & .47 & .31 & .32 & .86 & .48 & .36 & .45 & .48 & .32 & .34 & .86\\
 gpt3.5t      & .44 & .29 & .39 & .44 & .24 & .22 & .86 & .44 & .29 & .39 & .46 & .22 & .23 & .86\\
 gpt4         & .48 & .35 & .45 & .49 & .30 & .31 & .86 & .46 & .34 & .43 & .47 & .31 & .32 & .86\\
 T5b (ft)     & .49 & .36 & .46 & .47 & .36 & .29 & .89 & .52 & .37 & .48 & .48 & .38 & .28 & .90\\
 flanT5b (ft) & .49 & .36 & .47 & .47 & .43 & .23 & .90 & .50 & .37 & .47 & .48 & .41 & .29 & .91\\
 flanT5l (ft) & .49 & .36 & .47 & .45 & .42 & .28 & .90 & .51 & .38 & .47 & .48 & .42 & .29 & .91\\
 \hline
 \end{tabular}
}
 \caption{\textbf{Task 4} results: Automated Evaluation scores for model generated next actions for \textbf{actual and alternative} stories from the \textbf{test and dev splits}.  
 }
\label{tab:next-action-automated-metrics}
\vspace{-5mm}
\end{table*}

\begin{table*}[t]
\centering
\resizebox{.98\textwidth}{!}{
\small
\begin{tabular}[trim=0 .2in 0 0,  width=.95\linewidth]{p{0.28\linewidth} |p{0.12\linewidth} |p{0.65\linewidth} }
\hline \hline 
Story &  Model & Next Action\\
\hline 

\multirow{10}{=}{ \textbf{[Alternative] Story:} Jon and his father were going hunting in the woods. They saw a deer through the bushes. Trying to keep quiet, they inched towards it. Jon stepped on a branch, but the deer didn't flinch. Jon and his dad had deer for dinner that night.  
\textbf{Jon's Goal:Hunt and kill a deer.}
The next action involving Jon that is likely to happen after the last story sentence is  }
& Reference & \\
&fT5b (3-shot) &  Jon is going to be a doctor. and the reason for this action is he is going to be a doctor.\\
&fT5l (3-shot) & He ate the deer. \textbf{[repeats story details.]}\\
&fT5xl (3-shot) &  Jon ate the deer. \textbf{[repeats story details.]}\\
&fT5xxl (3-shot) &  Jon will be happy to have a deer for dinner. and the reason for this action is Jon and his father were going hunting in the woods.\\
&gpt3.5t (3-shot) &  Jon will learn to be more careful and quiet in the woods and the reason for this action is to avoid scaring the animals away. \textbf{[more appropriate for the actual story]}\\
&gpt4 (3-shot) & Clean his hunting gear. and the reason for this action is To prepare for the next hunting trip.\\
&T5-ft (0-shot) & he and his dad go hunting and the reason for this action is to get some food.\\
&fT5b-ft (0-shot) & They will go hunting again. and the reason for this action is They want to get more deer. \\
&fT5b-ft (3-shot) &  He will eat the deer. and the reason for this action is He wants to eat something he enjoyed. \textbf{[repeats story details ]}\\
\hline
\multirow{10}{=}{ 
\textbf{[Alternative] Story:} Lisa invited Tina and the girls over for a clambake. When the girls got there everything was fine. Yet all the sudden Rosie started raising her voice.  Lisa got very loud also. Lisa was so loud she asked Rosie to to sing louder. \textbf{Lisa' Goal: Have a nice dinner with her friends.} The next action involving Lisa that is likely to happen after the last story sentence is }
& Reference & Lisa will continue the bake with the rest of them. and the reason for the action is She must be having a good time if she asked Rosie to sing more.\\
&fT5b (3-shot)&  Lisa is going to be a good friend. and the reason for this action is Lisa is going to be a good friend.\\
&fT5l (3-shot) & Lisa yelled at Rosie. \textbf{[more appropriate for the actual story]}\\
&fT5xl (3-shot) & Lisa told Rosie to leave. \textbf{[more appropriate for the actual story]}\\
&fT5xxl (3-shot)&Lisa will ask Rosie to be quiet. and the reason for this action is Rosie is loud. \\
&gpt3.5t (3-shot)& Lisa apologizes to Rosie for getting loud and tries to calm the situation down. The reason for this action is to resolve the conflict and maintain a peaceful atmosphere at the clambake. \textbf{[more appropriate for the actual story]}\\
&gpt4 (3-shot)& Lisa will calm down and the reason for this action is that Rosie, the person causing her stress, has left. \textbf{[more appropriate for the actual story]}\\
&T5b-ft (0-shot)& she asks Rosie to sing louder and the reason for this action is to get her voice heard\\ 
&fT5b-ft (0-shot)& She will ask Rosie to sing louder. and the reason for this action is She wants to be heard.\\
&fT5b-ft (3-shot)&Lisa will ask Tina to sing louder. and the reason for this action is Lisa wants to hear her voice. \textbf{[This is not really a next action as it repeats story details.]}\\
\hline
\end{tabular}
}
\caption{\textbf{Task 3a} examples: Model generated Next Actions. Issues with generations described in square brackets. 
}
\label{tab:model-action-generations}
\vspace{-5mm}
\end{table*}

\begin{table*}[t]
 \centering
\resizebox{.98\textwidth}{!}{
 \begin{tabular}{|l|c|c|c|c|c|c|c||c|c|c|c|c|c|c|}
 \hline
       & \multicolumn{7}{|c||}{Actual Stories} & \multicolumn{7}{|c|}{Alternative Stories}\\
 Model & \multicolumn{3}{|c|}{Rouge} & Met- & \multicolumn{2}{|c|}{BLEU}&Bert & \multicolumn{3}{|c|}{Rouge} & Met- & \multicolumn{2}{|c|}{BLEU}&Bert\\ 
  & R1 & R2 & RL & eor & Cor. & Sen.&Score&R1 & R2 & RL & eor & Cor. & Sen.&Score\\ 
\hline
\hline
 T5-11b (3-shot)   & .11 & .02 & .10 & .19 & .01 & .02 & .86 & .11 & .02 & .09 & .19 & .00 & .02 & .86\\
 flanT5b (3-shot)  & .20 & .03 & .19 & .24 & .01 & .07 & .86 & .21 & .03 & .20 & .24 & .00 & .07 & .86 \\
 flanT5l (3-shot)  & .24 & .03 & .20 & .26 & .02 & .07 & .86 & .24 & .03 & .19 & .24 & .01 & .08 & .86 \\
 flanT5xl (3-shot) & .24 & .02 & .19 & .26 & .02 & .06 & .86 & .25 & .03 & .19 & .27 & .02 & .07 & .86\\
 flanT5xxl (3-shot)& .23 & .04 & .22 & .24 & .02 & .07 & .86 & .21 & .04 & .20 & .23 & .00 & .07 & .86\\
 gpt3.5t (3-shot)  & .22 & .05 & .19 & .28 & .03 & .06 & .86 & .19 & .04 & .17 & .26 & .02 & .06 & .86 \\
 gpt4  (3-shot)    & .23 & .04 & .19 & .29 & .03 & .07 & .86 & .23 & .04 & .18 & .29 & .03 & .06 & .86\\
 T5b (ft)          & .16 & .02 & .16 & .11 & .01 & .03 & .89 & .14 & .03 & .13 & .10 & .01 & .03 & .89\\
 flanT5b (ft)      & .16 & .03 & .15 & .09 & .01 & .04 & .90 & .14 & .04 & .12 & .09 & .03 & .04 & .90 \\
 flanT5l (ft)      & .17 & .03 & .16 & .11 & .01 & .04 & .90 & .15 & .04 & .14 & .10 & .03 & .04 & .90\\
\hline
\multicolumn{15}{|c|}{Dev Data Split}\\
\hline
 
 T5-11b (3-shot)   & .13 & .04 & .11 & .20 & .02 & .03 & .86 & .11 & .01 & .09 & .18 & .00 & .02 &.86\\
 flanT5b (3-shot)  & .22 & .03 & .19 & .20 & .01 & .07 & .86 & .19 & .01 & .17 & .20 & .00 & .06 &.86\\
 flanT5l (3-shot)  & .24 & .04 & .20 & .26 & .04 & .07 & .84 & .22 & .01 & .16 & .22 & .00 & .06 &.84\\
 flanT5xl (3-shot) & .22 & .04 & .18 & .23 & .04 & .08 & .85 & .21 & .01 & .16 & .24 & .00 & .06 &.85\\
 flanT5xxl(3-shot) & .22 & .05 & .19 & .22 & .05 & .08 & .86 & .19 & .02 & .17 & .20 & .00 & .06 &.86\\
 gpt3.5t  (3-shot) & .22 & .04 & .19 & .29 & .03 & .06 & .86 & .20 & .02 & .16 & .25 & .00 & .05 &.86\\
 gpt4     (3-shot) & .23 & .06 & .20 & .31 & .06 & .09 & .86 & .21 & .03 & .17 & .27 & .02 & .06 &.86\\
 T5b (ft)          & .18 & .09 & .17 & .14 & .06 & .08 & .89 & .13 & .02 & .12 & .08 & .01 & .04 &.88\\
 flanT5b (ft)      & .17 & .08 & .16 & .11 & .07 & .08 & .90 & .12 & .02 & .11 & .08 & .00 & .04 &.89\\
 flanT5l (ft)      & .13 & .05 & .12 & .09 & .03 & .05 & .89 & .14 & .03 & .13 & .08 & .01 & .03 & .89\\
\hline

\hline
 \end{tabular}
 }
 \caption{\textbf{Task 4} results: Automated and Human Evaluation scores for model generated Plans for goal achievement using \textbf{actual and alternative stories} from the \textbf{test and dev splits}. 
 }
\label{tab:future-plan-automated-metrics}
\end{table*}

\begin{table*}[t]
\centering
\resizebox{.98\textwidth}{!}{
\small
\begin{tabular}[trim=0 .2in 0 0,  width=.95\linewidth]{p{0.30\linewidth} |p{0.13\linewidth} |p{0.57\linewidth} }
\hline \hline 
Story &  Model & Plans\\
\hline 

\multirow{10}{=}{  
 \textbf{[Actual] Story:} Jared was tossing and turning all night. He kept hearing a buzz right next to his ear. Every time he tried to swat, the sound went away.  As he was drifting off, something landed right on top of his ear. He hit it and discovered that it was a now dead fly!  \textbf{Jared's Goal:} To sleep peacefully.
 \textbf{Next Action:} go to bed
 After the story and next action, Jared's plan to achieve the Goal is } 
 & Reference & he needs to get something to keep flies away.\\
&fT5b (3-shot)&  To get rid of the fly. \\
&fT5l (3-shot) & Sweat the fly. \textbf{[unrelated plan]}\\
&fT5xl (3-shot) & Jared will never swat at a fly again. \textbf{[unrelated plan]}\\
&fT5xxl (3-shot)&  Tossing and turning all night. \textbf{[not the best plan]}\\
& gpt3.5t (3-shot)&  Jared should try to keep his room more clean and ensure there are no insects or bugs flying around.  \\
&gpt4 (3-shot)&  Clean the room to prevent further disturbance from insects. \\
&T5b-ft (0-shot)& Jared needs to find a way to get a good night's sleep\\ 
&fT5b-ft (0-shot)& Jared should try to get a good night's sleep. \\
&fT5b-ft (3-shot)& Be more careful around the birds. \textbf{[unrelated plan]}\\
\hline 
\multirow{10}{=}{\textbf{[Actual] Story:} Ben had a new kite. Sam had a new kite as well. Sam's kite flew and Ben's didn't. Sam felt bad. Sam gave his kite to Ben and they were both happy.  \textbf{Ben's Goal:} Get his kite to fly.  \textbf{Next Action:} Ben will return the kite for another one. After the story and next action, Ben's plan to achieve the Goal is }
& Reference & No Plan.\\
& fT5b (3-shot)& to fly the kite.\\
& fT5l (3-shot)& to fly his kite more often\\
&fT5xl (3-shot)& Ben will try to fly his kite.\\
&ft5xxl (3-shot)&  to fly the kite\\
&gpt3.5t (3-shot)& Ben no longer needs a plan because he already achieved his goal of flying a kite. \textbf{[incorrect plan]}\\
&gpt4 (3-shot)& Learn more about kite flying to understand why his kite didn't take off. \textbf{[generic plan]}\\
& T5b-ft (0-shot)& Ben should get a new kite to fly.\\
& fT5b-ft (0-shot) & Ben will try to get his kite to fly again.\\
& fT5b-ft (3-shot) & Ben should have a new kite before he bought it. \textbf{[incorrect plan]}\\
\hline
\multirow{10}{=}{\textbf{[Alternative] Story:} Ben had a new kite. Sam had a new kite as well. Sam's kite flew and Ben's didn't.  Sam didn't care about Ben's kite. Sam kept flying his kite while Ben sat there upset.  \textbf{Ben's Goal:} Get his kite to fly.  \textbf{Next Action:} Ben will give up kite flying. After the story and next action, Ben's plan to achieve the Goal is }
& Reference & to find out what went wrong with Ben's kite and try again.\\
& fT5b (3-shot)& to fly the kite.\\
& fT5l (3-shot)& Sam's kite flew. \textbf{[unrelated plan]}\\
&fT5xl (3-shot)&  to try again \\
&ft5xxl (3-shot)&  to fly his kite\\
&gpt3.5t (3-shot)&  for Ben to ask Sam for tips on how to make his kite fly.\\
&gpt4 (3-shot)& Learn how to fly his kite properly.\\
& T5b-ft (0-shot)& Ben will give up kite flying.\\
& fT5b-ft (0-shot) & Ben will try to get his kite to fly.\\
& fT5b-ft (3-shot) & Ben should have a new kite.\\

\hline 
\hline 
\end{tabular}
}
\caption{\textbf{Task 4} examples: Model generated Plans. Issues with generations are described in
square brackets. }
\label{tab:model-plan-generations}
\vspace{-4mm}
\end{table*}

\begin{table*}[t]
\centering
\resizebox{.98\textwidth}{!}{
\small
\begin{tabular}[trim=0 .2in 0 0,  width=.95\linewidth]{p{0.23\linewidth} | p{0.77\linewidth}   }

\hline\hline

Task & Description of the Prompt using an example\\
\hline
T1. Goal Generation & Story: After work, Marie plopped on  ... Marie looked at her own kids and turned off the TV. \\
&Q: What primary goal did Marie hope to achieve through their actions in this story?   Very briefly describe the intention behind all or most of their actions in the story. \\
&A: The goal is \\
\hline
T2. Goal Transferability &Story: After work, Marie plopped on a chair to watch her shows. ... Marie rolled her eyes.\\
& Goal: wants to enjoy some ``me time.''\\
& Question: Does Marie intend to achieve the above goal in the story?\\
(Note: the story used &Select one of the options: \\
is an alternative story)&Yes\\
&No\\
&A: \\
\hline
T3. Next Action Generation & Story: After work, Marie plopped on a chair to watch her shows. ... Marie rolled her eyes. \\
& Goal:  wants to enjoy some ``me time.''\\
&Q: Very briefly describe  a specific action involving Marie that is likely to happen following the last story sentence and the reason for it.\\
&A: The next action is \\
\hline
T3b. Next Action Transfer &Story: After work, Marie plopped on a chair to watch her shows. ... Marie rolled her eyes.\\
& Goal:  wants to enjoy some ``me time.''\\
& Next Action: Marie plays with her kids.\\
& Q: Is Next Action the most likely action to happen after the last story sentence involving Marie?\\
(Note: the story used &Select one of the options: \\
is an alternative story)&Yes\\
&No\\
&Unsure\\
&A:\\
\hline
T3c. Explanation Transfer &Story: After work, Marie plopped on a chair to watch her shows. ... Marie rolled her eyes.\\
& Goal:  wants to enjoy some ``me time.''\\
& Next Action: Marie plays with her kids.\\
& Explanation: They want to get their mom's attention. \\
& Q: Does the Explanation provide the reason for why the Next Action is the most likely action following the last story sentence?\\
(Note the story used &Select one of the options: \\
is an alternative story)&Yes\\
&No\\
&A:\\
\hline

T4. Plan Generation &  Story: After work, Marie plopped on a chair to watch her shows. ... Marie rolled her eyes.\\
& Goal:  wants to enjoy some ``me time.''\\
& Next Action: Marie plays with her kids.\\
&Q: After the story and next action, what is Marie's plan to achieve the goal.\\
&A: The plan is\\
\hline
T5a. Achievement in story &Story: After work, Marie plopped on a chair to watch her shows. ... Marie rolled her eyes.\\
& Goal:  wants to enjoy some ``me time.''\\
& Q: Did Marie achieve their goal in the story?\\
&Select one of the options: \\
&Yes\\
&No\\
&Unsure\\
&A: \\
\hline
T5b. Achievement after story&Story: After work, Marie plopped on a chair to watch her shows. ... Marie rolled her eyes.\\
& Goal:  wants to enjoy some ``me time.''\\
& Next Action: Marie plays with her kids. \\
& Explanation: They want to get their mom's attention. \\
& Q: Does the next action help Marie achieve their goal?\\
&Select one of the options: \\
&Yes\\
&No\\
&Unsure\\
&A: \\\hline
T5c. Participant Satisfaction &Story: After work, Marie plopped on a chair to watch her shows. ... Marie rolled her eyes.\\
& Goal:  wants to enjoy some ``me time.''\\
& Next Action: Marie plays wtih her kids.\\
& Explanation: They want to get their mom's attention. \\
& Q: After the story and the next action is Marie satisfied with their goal achievement?\\
&Select one of the options: \\
&Yes\\
&No\\
&Unsure\\
&A: \\\hline
\hline
\end{tabular}
}
\caption{A detailed description of tasks and prompts.}
\label{tab:task-prompts}
\vspace{-4mm}
\end{table*}

\end{document}